\documentclass[letterpaper,10pt,times,conference]{IEEEtran}
\usepackage{lmodern}
\IEEEoverridecommandlockouts
\usepackage{cite}
\usepackage{amsmath,amssymb,amsfonts}
\usepackage[ruled,linesnumbered]{algorithm2e}
\usepackage[normalem]{ulem}
\usepackage{amssymb}
\usepackage{comment}
\usepackage{amsmath}
\usepackage{graphicx}
\usepackage{textcomp}
\usepackage{newtxtext}
\usepackage{hyperref}
\usepackage{multirow}
\usepackage{xcolor}
\usepackage{booktabs}
\usepackage{epsfig, subfigure, amsmath, amssymb, wrapfig}
\def\BibTeX{{\rm B\kern-.05em{\sc i\kern-.025em b}\kern-.08em
    T\kern-.1667em\lower.7ex\hbox{E}\kern-.125emX}}

\newcommand{\namet}{{\textsc{QPriv-VL}}\xspace}

\newcommand{\khalil}[1]{}
\author{
Md Khalid Syfullah and Alvi Ataur Khalil\\
Transformative Innovation for Trustworthy AI and Network Security (TITANS) Lab, \\ Computer Science, Southern Illinois University Carbondale, USA\\
\{mdkhalid.syfullah, a.khalil\}@siu.edu
\vspace{-10pt}}

\begin{document}
\title{Don't Send What You Don't Need: Question-Guided Token Pruning as a Privacy Defense for Vision-Language Models}

\maketitle
\thispagestyle{plain}
\pagestyle{plain}

\begin{abstract}
Visual Question Answering (VQA) with Vision-Language Models (VLMs) is increasingly deployed in privacy-sensitive and bandwidth-constrained settings. Distributed paradigms such as Federated Learning (FL), Split Learning (SL), and U-Shaped Split Learning (USL) keep raw data local; however, transmitting all visual tokens across the cut layer remains costly and vulnerable to privacy attacks. We propose \namet, a question-guided and privacy-aware token-pruning framework that operates uniformly across all three distributed setups and serves as a privacy defense by pruning visual tokens before they cross the cut layer using a combined measure of utility and sensitivity. At the core of \namet is a novel lightweight Dynamic Threshold Predictor (DTP), which, as an adaptive defense module, jointly predicts a per-sample prune ratio and a per-token retention mask in a single forward pass. DTP balances question relevance, estimated through a crossmodal comparison between each visual patch representation and the pooled question embedding, against a content-sensitivity signal derived from frozen features of DINOv2, a self-supervised vision transformer. DINOv2 enables DTP to identify visually sensitive regions while preserving question-relevant content via object-aware patch representations without labelled supervision. We evaluate \namet on three general-domain benchmarks (GQA, OKVQA, VQAv2), three medical benchmarks (SLAKE, VQA-RAD, Path-VQA), and four attack families (FSHA, FORA, iDLG, attribute-inference MIA). DTP matches or surpasses fixed-ratio pruning as a defense method while retaining substantially more tokens, reducing MIA attack success on VQA-RAD from $0.99$ to $0.76$--$0.79$, lowering FSHA and FORA reconstruction PSNR below the linear fixed-ratio trend, and preserving competitive VQA accuracy at roughly $40\%$ of the original token budget. A sensitivity exclusion ratio of $1.20 \pm 0.18$ shows that DTP preferentially removes privacy-sensitive patches, while explainability analysis confirms retention adapts to question semantics rather than generic saliency.
\end{abstract}


%%%%%%%%%%%%%%%%% Introduction %%%%%%%%%%%%
\section{Introduction}
\label{sec:introduction}
Visual Question Answering (VQA) is a prominent multimodal reasoning task where a model answers a natural language question based on the content of an image~\cite{antol2015vqa,hudson2019gqa}. Modern VQA systems are primarily powered by large Vision Language Models (VLMs) such as LLaVA~\cite{liu2023visual,liu2024improved}, BLIP2~\cite{li2023blip}, and Flamingo~\cite{alayrac2022flamingo}, which combine a Vision Transformer (ViT)~\cite{dosovitskiy2020image} encoder with a large language model via a cross-modal projector. While these models deliver high accuracy, their compute demands make on-device inference difficult.

Beyond raw compute, deployment is further complicated by privacy and throughput: many practical VQA applications operate in privacy-sensitive and bandwidth-constrained environments where transmitting raw visual data or deploying large models on-device is infeasible, such as multi-institutional medical VQA~\cite{lau2018dataset,liu2021slake,he2020pathvqa} and on-device assistive systems. These challenges have motivated distributed training paradigms, notably Federated Learning (FL)~\cite{mcmahan2017communication} and Split Learning (SL)~\cite{gupta2018distributed}. FL preserves data locality by aggregating model updates without sharing raw data, whereas SL partitions the model between client and server, transmitting only intermediate activations across a cut layer. U-Shaped Split Learning (USL) extends SL with a second cut layer, keeping both labels and the question head on the client side~\cite{vepakomma2018split}.

While suited for VLM-based VQA, these paradigms introduce three key challenges. First, communication overhead is substantial: a CLIP-ViT-L/14 encoder at $336{\times}336$ resolution produces $576$ visual tokens with $1024$-dimensional embeddings, making transmission across the SL cut prohibitively expensive~\cite{eshratifar2019bottlenet,hassan2025spikebottlenet}. Second, intermediate activations and gradient updates are vulnerable to feature-inversion, gradient-leakage, and membership-inference attacks~\cite{pasquini2021unleashing,xu2024stealthy,zhao2020idlg,shokri2017membership}, against which existing activation-level defenses reduce utility while ignoring that many transmitted tokens carry no question-relevant information. Third, VLM analyses show that only a small subset of visual tokens contributes after the early language-model layers~\cite{chen2024image,zhang2024sparsevlm}, meaning transmitting the full prefix inflates both communication cost and attack surface.

These challenges motivate a simple observation: only a subset of visual tokens is necessary to cross the cut layer. Selecting task-relevant and privacy-preserving tokens before transmission removes redundancy, shrinks the attack surface, and preserves semantically meaningful information. Recent question-guided and adaptive token-pruning approaches for VLMs reduce the visual prefix by two-thirds or more~\cite{chen2024image,zhang2024sparsevlm,ye2025atp,li2025qg,guo2025crop,xing2024pyramiddrop,yang2025visionzip}. However, these methods are evaluated as centralized inference accelerators and, to our knowledge, remain unexplored as privacy defenses or jointly trained with content-sensitivity signals. Existing SL compression approaches are largely task-agnostic and do not exploit the client-side question features.

To bridge this gap, we introduce \namet, a \underline{\textbf{Q}}uestion-guided and \underline{\textbf{Priv}}acy-aware token-pruning framework for \underline{\textbf{V}}ision-\underline{\textbf{L}}anguage models, built on LLaVA-1.5-7B~\cite{liu2024improved}, which combines a CLIP-ViT visual encoder with a Llama-2 language backbone via a lightweight Multi Layer Perception (MLP) projector. \namet operates as a unified client-side pruning framework for Centralized Training (CL), FL, SL, and USL, using the same decision mechanism regardless of cut placement. A single checkpoint shortens the visual prefix, reduces communication cost across the cut, and limits the information exposed to curious or malicious servers. The framework centers on a lightweight client-side Dynamic Threshold Predictor (DTP), where one forward pass produces a per-sample pruning ratio (how many tokens leave the client) and a per-token keep-mask (which tokens are retained). Our contribution in this work is four-fold:
\begin{itemize}
\item We introduce a lightweight, privacy-aware DTP module that decides how many and which visual tokens to keep, jointly weighing question relevance against a learned content-sensitivity signal. DTP generalizes across general and medical domains without per-domain retraining.
\item We present a unified implementation of \namet on LLaVA-1.5-7B that supports both fixed and dynamic thresholds, letting the same client-side module act as a communication-efficiency layer and a privacy defense under distributed settings.
\item We provide a systematic privacy evaluation of \namet, showing robustness against four representative attack families: FSHA~\cite{pasquini2021unleashing}, FORA~\cite{xu2024stealthy}, iDLG~\cite{zhao2020idlg}, and MIA~\cite{shokri2017membership}, on both general-domain and medical VQA benchmarks.
\item We investigate the utilised VLM model's internal decision process via transformer-based attention visualization~\cite{chefer2021transformer}, using case studies with in and out-of-context queries.
\end{itemize}
This work answers four research questions:
\begin{itemize}
\item \textbf{RQ1:} Does \namet simultaneously reduce communication cost, shrink the attack surface, and preserve accuracy across all four learning settings?
\item \textbf{RQ2:} Does a privacy-aware dynamic threshold predictor yield a better accuracy/communication/privacy Pareto trade-off than fixed-ratio or training-free pruning?
\item \textbf{RQ3:} Does \namet specifically remove tokens that are both reconstructable and privacy-relevant, rather than merely degrading performance?
\item \textbf{RQ4:} Can its effectiveness be verified through explainability analysis?
\end{itemize}

The remainder of this paper is organized as follows. Section~\ref{sec:background} presents the background. Section~\ref{sec:literature_review} discusses related work. Section~\ref{sec:proposed_framework} introduces the proposed framework. Section~\ref{sec:methodology} outlines the methodology. Section~\ref{sec:experimental-results} reports the experimental findings and section~\ref{sec:conclusion} concludes the paper.
%%%%%%%%%%%%%%%%%%%%%%%%%%%%%%%%%%%%%%%%%%%%

%%%%%%%%%%%%%%%%% Background %%%%%%%%%%%%%
\section{Background}
\label{sec:background}
In this section, we discuss preliminary concepts to facilitate an understanding of distributed learning, attacks, and the positioning of our proposed \namet framework.

%==============================================
\subsection{VQA and VLMs}
\label{subsec:bg-vqa}
VQA takes an image together with a natural-language question and produces a free-form textual answer~\cite{antol2015vqa}. Modern systems such as LLaVA~\cite{liu2023visual,liu2024improved}, BLIP-2~\cite{li2023blip}, and Flamingo~\cite{alayrac2022flamingo} realise this mapping with a VLM that pairs a visual encoder with a language model through a small projection module, such that visual features are translated into a form the language model can attend to alongside the question tokens. We use LLaVA-1.5-7B~\cite{liu2024improved} as our reference architecture because its CLIP-ViT-L/14 encoder, MLP projector, and Vicuna language model expose a clean interface between modalities and have an established literature on visual-token pruning~\cite{chen2024image,zhang2024sparsevlm,ye2025atp,shang2025llava}.

%==============================================
\subsection{Self-Supervised Visual Representations}
\label{subsec:bg-dinov2}
Self-supervised visual representation learning trains a ViT on a large unlabelled image corpus by predicting one view of an image from another. The resulting per-patch features carry strong semantic locality and transfer well across visual domains, including medical imagery, without retraining. We adopt DINOv2~\cite{oquab2023dinov2} as the self-supervised backbone behind our sensitivity branch, since it offers high-quality patch features at a small computational cost and remains frozen.

%==============================================
\begin{figure}[t]
\centering
\includegraphics[width=\columnwidth]{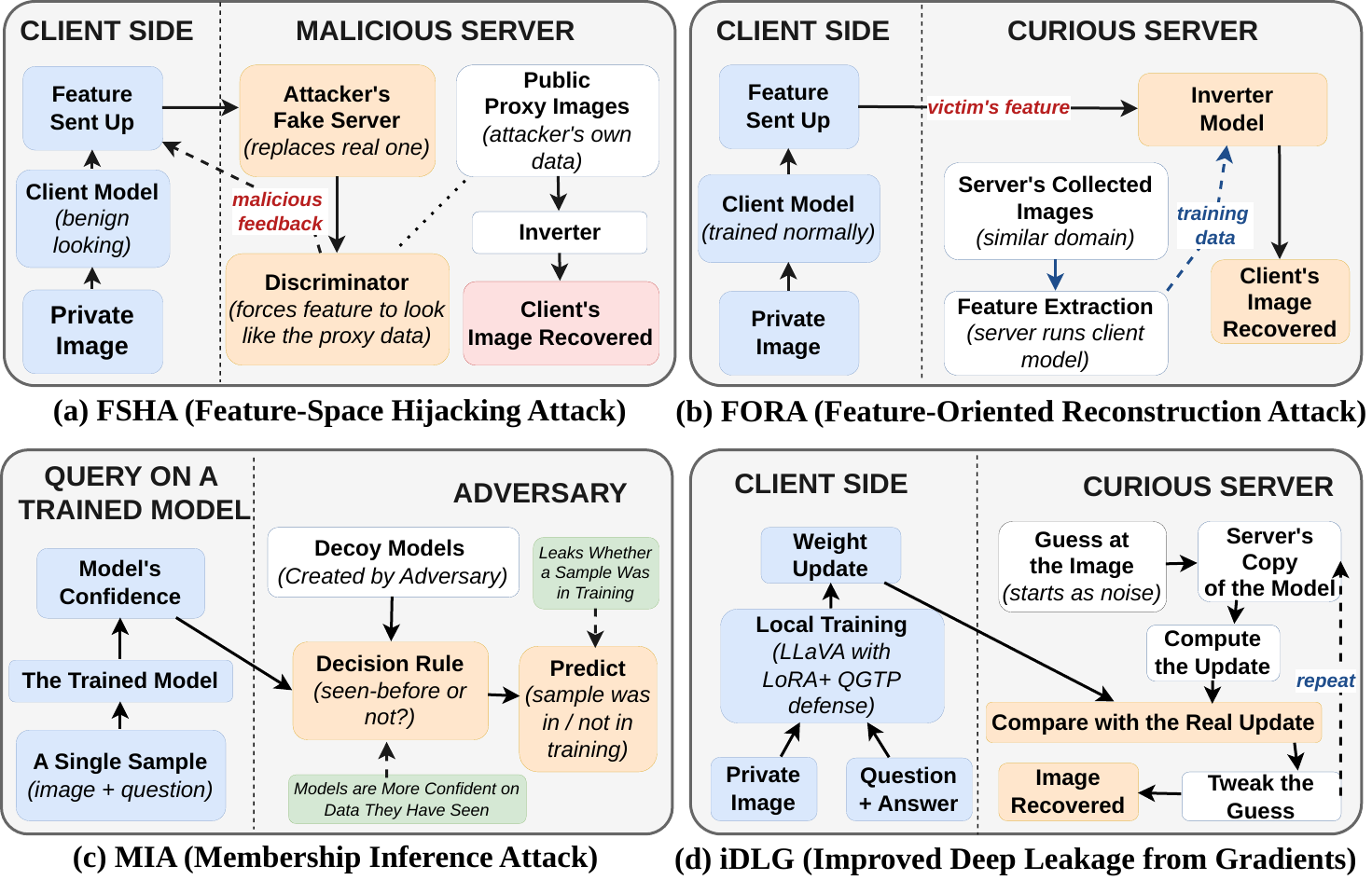}
\vspace{-20pt}
\caption{Overview of the four attacks considered in this work.}
\vspace{-15pt}
\label{fig:attacks}
\end{figure}
%==============================================

%==============================================
\subsection{Training Paradigms}
\label{subsec:bg-distributed}
We compare four training paradigms differing in how data and the model are partitioned between client and server.
\paragraph{CL} The classical setting where a single trainer holds the entire dataset and updates a single copy of the model. CL serves as the no-distribution upper bound: nothing crosses a network boundary, eliminating any leakage channel, though raw data must be aggregated in one place.
\paragraph{FL~\cite{mcmahan2017communication}} Each client keeps its data local and trains a local copy. Clients periodically send parameter updates to a server, which aggregates them and broadcasts the result. FL preserves data locality; transmitted updates can still leak information through gradient-inversion attacks such as DLG and iDLG~\cite{zhu2019deep,zhao2020idlg}.
\paragraph{SL~\cite{gupta2018distributed}} The model is partitioned at a single cut layer. The client owns the lower part and the input data; the server owns the upper part and the loss computation. Only the intermediate activations at the cut and the back-flowing gradients are exchanged. The smashed activations become the primary target for feature-inversion attacks~\cite{pasquini2021unleashing,xu2024stealthy}.
\paragraph{USL~\cite{vepakomma2018split}} USL extends SL with a second cut so both ends of the network remain on the client side: the client retains the embedding layers, the language-model head, and the labels, while the server operates only on the middle block. This topology removes several leakage channels at the cost of an extra round of communication per step.

%==============================================
\subsection{Attacks on VLMs}
\label{subsec:bg-attacks}
A central privacy concern in distributed learning is that an honest-but-curious or malicious server can recover the input, its label, or a sensitive attribute from exchanged information (e.g., parameter updates in FL or smashed activations and back-flowing gradients at the cut layer(s) in SL and USL). We adopt four attacks spanning the threat space (Figure~\ref{fig:attacks}).

\paragraph{FSHA} Pasquini et al.~\cite{pasquini2021unleashing} introduce an active attack where an attacker-controlled module replaces the legitimate server and hijacks training: a discriminator on the attacker side covertly aligns the client's smashed representations with embeddings of the attacker's public proxy images, while an inverter trained jointly against this aligned space reconstructs the client's private input (Figure~\ref{fig:attacks}(a)). The attack defeats distance-correlation defenses because the alignment objective collapses the distances those defenses regularise.

\paragraph{FORA} Xu et al.~\cite{xu2024stealthy} propose a passive reconstruction attack in which an honest-but-curious server collects smashed activations from a trained client model and trains an inverter on a same-domain public auxiliary set, using its copy of the client encoder to generate matched feature/image pairs (Figure~\ref{fig:attacks}(b)). Because the inverter is fit against unpruned, undefended representations, FORA acts as an oracle attacker that bounds what any adversary could recover from the deployed system.

\paragraph{MIA} Following Shokri et al.~\cite{shokri2017membership}, an adversary trains shadow models on data from the target's distribution and uses their confidence patterns to build a binary classifier that predicts whether a candidate sample was in the training set (Figure~\ref{fig:attacks}(c)). The signal exploited is the well-known tendency of models to be more confident on samples they have seen during training, providing a leakage channel orthogonal to reconstruction.

\paragraph{iDLG} DLG~\cite{zhu2019deep} optimises a dummy input whose synthetic gradient matches an observed gradient update, iteratively refining a noise initialisation until the reconstructed gradient agrees with the intercepted one and the dummy input converges to the private image; iDLG~\cite{zhao2020idlg} additionally recovers the label analytically from the sign of the final-layer gradient. The attack applies to FL on transmitted parameter updates and to SL by inverting the gradient flowing back across the cut (Figure~\ref{fig:attacks}(d)).
%%%%%%%%%%%%%%%%%%%%%%%%%%%%%%%%%%%%%%%%%%%%

%%%%%%%%%%%%%%%%% Literature Review %%%%%%%%%%%%%
\section{Literature Review}
\label{sec:literature_review}
\namet sits at the intersection of two largely separate lines: question-guided token pruning for VLMs targeting centralised inference efficiency, and privacy-preserving distributed deep learning, whose defenses operate on activations or gradients without semantic token control.

Question-guided and adaptive token pruning is directly related. FastV~\cite{chen2024image} prunes low-attention tokens in deeper LLM layers; SparseVLM~\cite{zhang2024sparsevlm} and ATP-LLaVA~\cite{ye2025atp} add text-conditioned per-sample budgets; QG-VTC~\cite{li2025qg} learns question-conditioned token scores; and CROP~\cite{guo2025crop} adapts retention to query-relevant regions. PyramidDrop~\cite{xing2024pyramiddrop} and VisionZip~\cite{yang2025visionzip} interleave dropping with progressive selection across LLM depths, while LLaVA-PruMerge~\cite{shang2025llava} combines outlier-based selection with attention-guided merging. These methods show two-thirds or more of the visual prefix is redundant for VQA, yet place the decision inside the LLM and serve only as centralised inference accelerators. None positions it before a network boundary, co-trains it with a content-sensitivity signal, or measures its effect on inversion-based or membership-inference attackers.

Privacy in distributed deep learning has been extensively studied for CNN-style architectures. Feature-inversion attacks~\cite{pasquini2021unleashing,xu2024stealthy} reconstruct private inputs from smashed activations, and gradient-leakage attacks~\cite{zhu2019deep,zhao2020idlg} invert gradient updates; defenses such as NoPeek~\cite{vepakomma2020nopeek}, DP-CutMixSL~\cite{baek2022visual}, and PATROL~\cite{ding2024patrol} mitigate leakage via distance-correlation regularisation, sample mixing, or channel pruning. Communication-efficient SL methods such as BottleNet~\cite{eshratifar2019bottlenet} and ADC~\cite{alvetreti2025communication} cut activation bandwidth via feature compression or attention-driven dropping, though without question conditioning. For multimodal systems, FedMultimodal~\cite{feng2023fedmultimodal} and pFL-MedVQA~\cite{zhu2024client} focus on FL parameter aggregation, while BiCSL~\cite{sun2024bidirectional} combines SL and FL for VQA under a contrastive privacy objective without token pruning, leaving the visual prefix and its reconstructible surface untouched. Overall, existing defenses target CNN-style activations or gradients without exploiting question semantics, so multimodal extensions still transmit the full visual prefix and leave its reconstructible surface largely unaddressed.

Self-supervised vision transformers DINO~\cite{caron2021emerging} and DINOv2~\cite{oquab2023dinov2} learn label-free, object-aware patch representations transferable across natural and medical domains. For faithfulness evaluation, transformer-aware attention propagation~\cite{chefer2021transformer} and deletion-based probes~\cite{petsiuk2018rise,deyoung2020eraser} are commonly used in VLM explainability analysis. Together, these tools provide a label-free signal for identifying privacy-relevant regions and a principled means of verifying that pruning decisions reflect question-driven reasoning rather than generic saliency.

%==============================================
\begin{figure*}[t]
  \centering
  \includegraphics[width=0.8\linewidth]{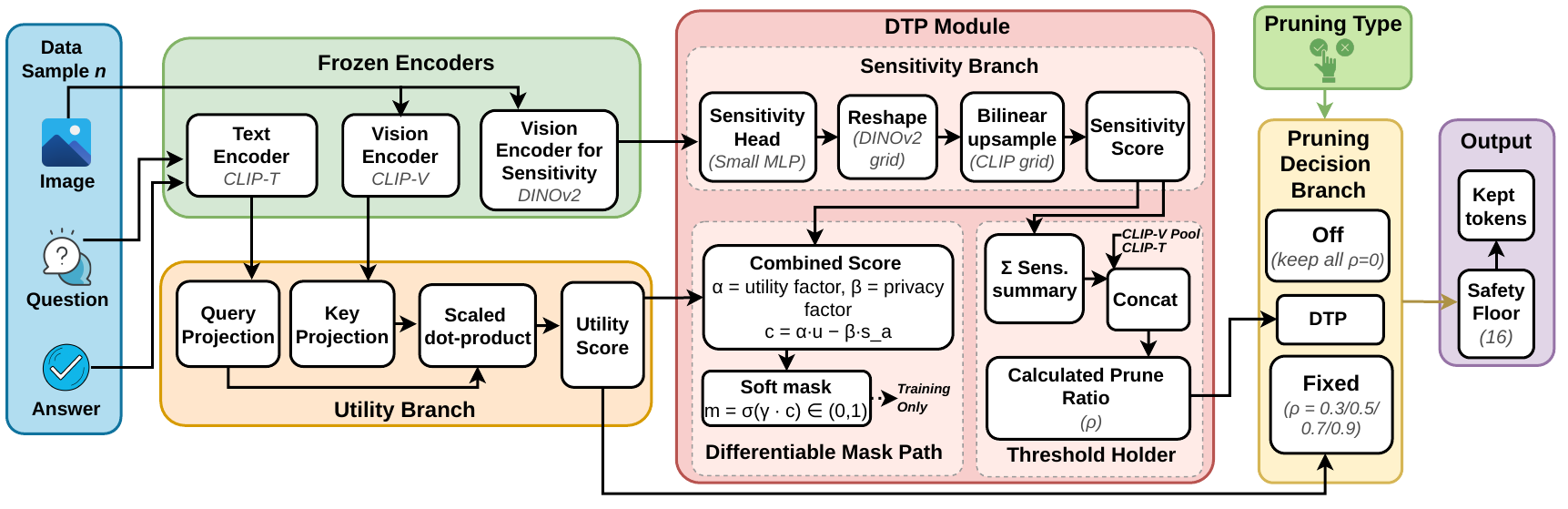}
  \vspace{-10pt}
  \caption{The proposed \namet framework.}
  \vspace{-15pt}
  \label{fig:pipeline}
\end{figure*}
%==============================================

Prior work studies question-guided VLM pruning, privacy-preserving SL/FL, and self-supervised sensitivity modelling separately. VLM pruning methods focus on inference cost in centralised settings, while privacy defenses protect CNN-style features or gradients without controlling semantic tokens. Distributed multimodal methods transmit the full visual sequence. \namet addresses this gap by combining question-guided token pruning with learned per-sample ratios and token-level keep-masks, integrating DINOv2-based patch sensitivity, evaluating a unified policy across CL, FL, SL, and USL, and analysing robustness against four attacks.
%%%%%%%%%%%%%%%%%%%%%%%%%%%%%%%%%%%%%%%%%%%%

%%%%%%%%%%%%%%%%% Framework %%%%%%%%%%%%%
\section{Proposed Framework}
\label{sec:proposed_framework}
The \namet framework performs Question-Guided Token Pruning (QGTP) and sits between the frozen vision-language encoders and the LLaVA multi-modal projector (Figure~\ref{fig:pipeline}). Given an image-question pair, it determines both the quantity and selection of visual patches to pass forward, blocking question-irrelevant or privacy-sensitive content from leaving the client. This same framework operates unchanged across CL, FL, SL, and USL, requiring no architectural adjustments.
%==============================================
\subsection{Frozen Encoders}
\label{subsec:fw-encoders}
Three pre-trained backbones produce fixed feature representations. The CLIP vision encoder (CLIP-V) turns the image into a grid of patch tokens and a pooled image summary. The CLIP text encoder (CLIP-T) turns the question into a pooled vector. A self-supervised DINOv2 encoder produces a grid of patch features emphasising object-level structure independent of text supervision; these feed the privacy-sensitivity signal.
%==============================================
\subsection{Utility Branch}
\label{subsec:fw-utility}
The utility branch assigns a question-relevance score to each CLIP patch. The pooled question embedding and CLIP patch tokens are projected into a shared low-dimensional space via query and key projections. A scaled dot-product attention followed by a sigmoid then produces a utility score per patch, emphasising regions most relevant to the question.
%==============================================
\subsection{DTP Module}
\label{subsec:fw-dtp}
The DTP module fuses utility with privacy sensitivity, predicts the per-sample prune ratio, and is the only learnable component of the pipeline. It groups three sub-components:
\paragraph{Sensitivity Branch}
DINOv2 patch features are passed through a small per-patch MLP (the sensitivity head), reshaped onto the DINOv2 grid, and bilinearly upsampled onto the CLIP grid. Every CLIP patch therefore carries a utility and a sensitivity score on the same spatial layout.
\paragraph{Differentiable Mask Path}
A combined score is computed by subtracting weighted sensitivity from weighted utility, followed by a sigmoid with a learnable sharpness parameter to generate a soft per-patch keep-mask in $(0,1)$. During training, this mask acts as a differentiable distillation target; at inference it is replaced with a discrete top-$k$ token selection.
\paragraph{Threshold Holder}
A compact summary of the sensitivity field (mean, maximum, fraction above a threshold) is concatenated with the pooled image and question descriptors, and a small MLP outputs the per-sample prune ratio, the only quantity varying per sample at inference.
%==============================================
\subsection{Pruning Decision Branch}
\label{subsec:fw-pruning}
The pruning decision branch selects one of three policies for each forward pass: \textsc{off} (no pruning, $\rho{=}0$, used as the no-defense baseline), \emph{Fixed} (a manually chosen $\rho \in \{0.3, 0.5, 0.7, 0.9\}$), or \emph{DTP}. The chosen ratio drives a top-$k$ selection over the utility scores produced by the utility branch.
%==============================================
\subsection{Output}
\label{subsec:fw-output}
The retained patches are assembled into a variable-length visual prefix, which is passed through LLaVA's multi-modal projector and combined with the question token embeddings before being processed by the language model. To preserve sufficient visual information under aggressive pruning, a safety threshold of $k_{\min}{=}16$ tokens is enforced, ensuring the answer generator always receives a minimum amount of image context. From the LLM's perspective, the primary modification is the shorter, more selectively curated visual prefix. During downstream fine-tuning, only the language model is adapted using LoRA adapters~\cite{hu2022lora}, the pruning module remains fixed.
%%%%%%%%%%%%%%%%%%%%%%%%%%%%%%%%%%%%%%%%%%%%

%%%%%%%%%%%%%%%%% Methodology %%%%%%%%%%%%%
\section{Methodology}
\label{sec:methodology}
This section gives the technical specification of every component including the \namet framework, the DTP network and its preparation pipeline, the four training configurations, and the robustness and privacy probes used in the evaluation. Table~\ref{tab:notation} summarises the notation used throughout the paper.

%==============================================
\begin{table}[t]
\centering
\caption{Summary of notation used throughout the paper.}
\label{tab:notation}
\footnotesize
\setlength{\tabcolsep}{4pt}
\scriptsize
\begin{tabular}{|c|l|}
\hline
\textbf{Symbol} & \textbf{Description} \\
\hline
$I_b, q_b, a^\star_b$ & Image, question, and answer for sample $b$ \\
\hline
$N$ & Number of CLIP-V patch tokens ($N{=}576$, $24{\times}24$ grid) \\
\hline
$V_b$ & CLIP-V patch tokens, $V_b \in \mathbb{R}^{N \times D_v}$, $D_v{=}1024$ \\
\hline
$v_b^{\text{pool}}$ & CLIP-V pooled image feature, $v_b^{\text{pool}} \in \mathbb{R}^{D_v}$ \\
\hline
$t_b$ & CLIP-T pooled question embedding, $t_b \in \mathbb{R}^{D_t}$, $D_t{=}768$ \\
\hline
$d_b$ & DINOv2 patch tokens, $d_b \in \mathbb{R}^{256 \times D_d}$, $D_d{=}384$ \\
\hline
$\psi$ & LLaVA multi-modal projector \\
\hline
$W_q, W_k$ & Scorer projections to shared dim $P{=}256$ \\
\hline
$u_{b,n}$ & Question-conditioned utility score for patch $n$ \\
\hline
$H_{\text{sens}}$ & Per-patch sensitivity MLP \\
\hline
$s^{\text{dino}}_{b,m}$ & Per-patch sensitivity on DINOv2 grid \\
\hline
$s^{\text{aligned}}_b$ & Sensitivity upsampled to the CLIP $24{\times}24$ grid \\
\hline
$c_{b,n}$ & Combined score $\alpha u_{b,n} - \beta s^{\text{aligned}}_{b,n}$ \\
\hline
$m_{b,n}$ & Soft keep-mask, $m_{b,n} = \sigma(\gamma c_{b,n})$ \\
\hline
$\alpha, \beta$ & Utility/sensitivity weights (buffers, $\alpha{=}\beta{=}1$) \\
\hline
$\gamma$ & Learnable mask sharpness (initialised to $3.0$) \\
\hline
$\rho_b, \hat\rho_b$ & True/predicted per-sample prune ratio \\
\hline
$\rho^\star_b$ & Oracle prune ratio from LLaVA-in-the-loop search \\
\hline
$k_b, k_{\min}$ & Number of retained tokens; safety floor ($k_{\min}{=}16$) \\
\hline
$\mathcal{S}_b$ & Index set of retained patches \\
\hline
$\mathrm{MLP}_\tau$ & Threshold predictor producing $\hat\rho_b$ \\
\hline
$\theta$ & DTP parameters $\{W_q, W_k, H_{\text{sens}}, \mathrm{MLP}_\tau, \gamma\}$ \\
\hline
$\mathcal{L}_{\text{sens}}, \mathcal{L}_\tau$ & Teacher sensitivity and threshold losses\\
\hline
$\mathcal{L}_\rho, \mathcal{L}_m, \mathcal{L}_{\text{KD}}$ & Student ratio, mask, and combined KD losses \\
\hline
$\mathcal{D}_{\text{vqa}}, \mathcal{D}_{\text{priv}}$ & VQA and auxiliary privacy training sets \\
\hline
$L$ & Number of LLM decoder layers ($L{=}32$) \\
\hline
$C, C_A, C_B$ & Split cut layers (SL: $C{=}16$; USL: $C_A{=}8$, $C_B{=}24$) \\
\hline
$N_C, T_l$ & Number of clients and local steps per round \\
\hline
$\mathcal{G}$ & Oracle prune-ratio search grid \\
\hline
\end{tabular}
\vspace{-10pt}
\end{table}
%==============================================

%==============================================
\subsection{DTP Preparation}
\label{sec:dtptraining}
The DTP preparation pipeline (Algorithm~\ref{alg:dtp}) has three stages: (i) feature pre-computation, (ii) oracle prune-ratio search, and (iii) teacher-student Knowledge Distillation. Training uses six VQA benchmarks $\mathcal{D}_{\text{vqa}}$ spanning natural images (VQAv2~\cite{goyal2017making}, GQA~\cite{hudson2019gqa}, OK-VQA~\cite{schwenk2022okvqa}) and medical imaging (SLAKE~\cite{liu2021slake}, VQA-RAD~\cite{lau2018dataset}, PathVQA~\cite{he2020pathvqa}), each capped at $3{,}000$ samples. The sensitivity branch $H_{\text{sens}}$ is trained on a separate auxiliary set $\mathcal{D}_{\text{priv}}$ with image-level privacy labels drawn from CelebA, NIH Chest X-Ray, EuroSAT (positive bags) and MNIST, CIFAR-10 (negative bags); these labels are used \emph{only} for $H_{\text{sens}}$ and never enter the VQA loop.

\paragraph{Feature pre-computation}
For each sample $(I_b, q_b, a^\star_b) \in \mathcal{D}_{\text{vqa}}$, the three frozen encoders are evaluated once and the tuple $(V_b, v_b^{\text{pool}}, t_b, d_b)$ is stored in half precision for reuse in later stages.

\paragraph{Oracle prune-ratio search}
For each sample we identify the smallest prune ratio under which LLaVA-1.5-7B still recovers the gold answer. Patches are ranked by a fixed, randomly initialised scorer $\tilde u$. For each $\rho \in \mathcal{G} = \{0.10, 0.20, 0.30, 0.50, 0.70\}$ in ascending order, the top-$\lceil (1-\rho) N \rceil$ patches are passed through LLaVA's projector $\psi$ and language model; the prediction is matched against $a^\star_b$ under a first-token substring criterion. The smallest matching $\rho$ defines $\rho^\star_b$; if no value matches, $\rho^\star_b = 1.0$. The search is capped at $500$ samples per dataset.

\paragraph{Teacher-student knowledge distillation}This stage is the largest and consists of three parts: teacher training, the rationale for distillation, and student distillation.

\noindent\textbf{Teacher training:} The teacher shares DTP's forward equations, training only $H_{\text{sens}}$ and $\mathrm{MLP}_\tau$; $(W_q, W_k)$ remain at random initialisation and mask sharpness is fixed at $\gamma=3$. The two heads are trained sequentially. With image-level labels $y_b \in \{0,1\}$ from $\mathcal{D}_{\text{priv}}$ and $\hat y_b = \max_m s^{\text{dino}}_{b,m}$, $H_{\text{sens}}$ minimises a binary cross-entropy loss for $10$ epochs (AdamW, lr $3{\times}10^{-4}$, weight decay $10^{-4}$):
\begin{equation}
\mathcal{L}_{\text{sens}} = -\tfrac{1}{B}\sum_{b}\bigl[\, y_b \log \hat y_b + (1-y_b)\log(1-\hat y_b)\,\bigr].
\label{eq:loss-sens}
\end{equation}
With $H_{\text{sens}}$ frozen, $\mathrm{MLP}_\tau$ then regresses the oracle ratio on $\mathcal{D}_{\text{vqa}}$ for $20$ epochs (AdamW, lr $10^{-4}$, weight decay $10^{-2}$, dropout $0.1$, batch size $64$):
\begin{equation}
\mathcal{L}_{\tau} = \tfrac{1}{B}\sum_{b}\bigl(\hat\rho_b - \rho^\star_b\bigr)^2.
\label{eq:loss-tau}
\end{equation}

\noindent\textbf{Rationale:} The oracle ratio $\rho^\star_b$ is costly because it requires a per-sample LLaVA-based grid search over $|\mathcal{G}|$ candidates and is discrete since it can only take values on the search grid. In addition, the privacy-supervised teacher head $H^T_{\text{sens}}$ is trained on a separate auxiliary set $\mathcal{D}_{\text{priv}}$ that is not available to the student. As a result, neither signal can directly serve as the sole supervision for a unified differentiable student that must jointly predict $\hat\rho_b$ and a soft keep-mask $m_b$ in a single forward pass. Knowledge distillation addresses this by transferring both signals through end-to-end training: the soft teacher outputs provide smoother supervision beyond the discrete oracle ratios, while the hard oracle targets regularise the model against teacher miscalibration and oracle saturation cases.

\noindent\textbf{Student distillation:} DTP is initialised from the teacher's $H_{\text{sens}}$ and $\mathrm{MLP}_\tau$, with $W_q, W_k$ random and $\gamma{=}3.0$, and is trained end-to-end against both the teacher's outputs $(\rho^T_b, m^T_b)$ and the hard oracle $\rho^\star_b$:
\begin{align}
\mathcal{L}_\rho &= \tfrac{1}{2}(\hat\rho_b - \rho^T_b)^2 + \tfrac{1}{2}(\hat\rho_b - \rho^\star_b)^2,
\label{eq:loss-rho}\\
\mathcal{L}_m &= \mathrm{BCE}(m_b, m^T_b),
\label{eq:loss-m}\\
\mathcal{L}_{\text{KD}} &= \lambda_\rho\,\mathcal{L}_\rho + \lambda_m\,\mathcal{L}_m,
\label{eq:loss-kd}
\end{align}
with $\lambda_\rho = \lambda_m = 1$. The mask term drives $(W_q, W_k)$ to mimic the teacher's saliency. Training runs for $20$ epochs (AdamW, lr $3{\times}10^{-4}$ with cosine annealing, batch size $128$).

%==============================================
\begin{algorithm}[t]
\DontPrintSemicolon
\SetKwInOut{KwIn}{Input}
\SetKwInOut{KwOut}{Output}
\KwIn{$\mathcal{D}_{\text{vqa}}$, $\mathcal{D}_{\text{priv}}$, prune-ratio grid $\mathcal{G}$}
\KwOut{DTP param. $\theta = \{W_q, W_k, H_{\text{sens}}, \mathrm{MLP}_\tau, \gamma\}$}
\BlankLine
\ForEach{$(I_b, q_b) \in \mathcal{D}_{\text{vqa}}$}{
   $(V_b, v_b^{\text{pool}}, t_b, d_b) \gets \mathrm{Enc}(I_b, q_b)$\;
}
\BlankLine
\ForEach{$b \in \mathcal{D}_{\text{vqa}}$}{
   $\rho^\star_b \gets 1.0$\;
   \ForEach{$\rho \in \mathcal{G}$ in ascending order}{
      $k \gets \lceil(1-\rho)\,N\rceil$\;
      $\mathcal{S} \gets \mathrm{top\text{-}}k\bigl(\tilde u(V_b, t_b)\bigr)$\;
      \If{$\mathrm{LLaVA}\bigl(\psi(V_b[\mathcal{S}]), q_b\bigr) \approx a^\star_b$}{
         $\rho^\star_b \gets \rho$\;
         \textbf{break}\;
      }
   }
}
\BlankLine
$H_{\text{sens}}^T \gets \arg\min_{H_{\text{sens}}} \mathcal{L}_{\text{sens}}(\mathcal{D}_{\text{priv}})$\;
$\mathrm{MLP}_\tau^T \gets \arg\min_{\mathrm{MLP}_\tau} \mathcal{L}_{\tau}(\mathcal{D}_{\text{vqa}})$\;
$\theta \gets \mathrm{init}\bigl(H_{\text{sens}}^T,\, \mathrm{MLP}_\tau^T,\, \mathrm{rand}(W_q, W_k),\, \gamma{=}3.0\bigr)$\;
$\theta \gets \arg\min_{\theta} \mathcal{L}_{\text{KD}}(\mathcal{D}_{\text{vqa}})$\;
\Return $\theta$\;
\caption{DTP preparation pipeline.}
\label{alg:dtp}
\end{algorithm}
%==============================================

%==============================================
\subsection{\namet Framework Architecture}
\label{sec:qgtp}
%==============================================
\begin{figure*}[t]
  \centering
  \includegraphics[width=\linewidth]{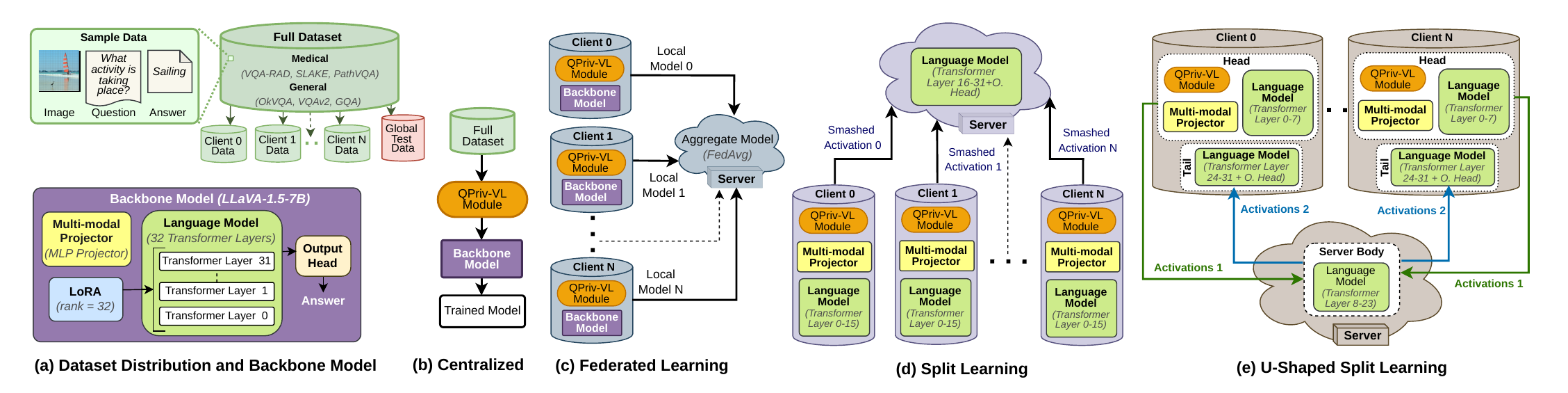}
  \vspace{-18pt}
  \caption{Dataset distribution, backbone model, and system architecture across the four training configurations.}
  \vspace{-15pt}
  \label{fig:sysarc}
\end{figure*}
%==============================================

We summarise the dimensions and equations defining the \namet framework from Section~\ref{sec:proposed_framework}. CLIP-V produces $V_b \in \mathbb{R}^{N \times D_v}$ from its penultimate layer ($N{=}576$, $D_v{=}1024$), along with pooled features $v_b^{\text{pool}} \in \mathbb{R}^{D_v}$. CLIP-T~\cite{radford2021learning} produces $t_b \in \mathbb{R}^{D_t}$ with $D_t{=}768$. DINOv2~\cite{oquab2023dinov2} produces $d_b \in \mathbb{R}^{256 \times D_d}$ on a $16{\times}16$ grid with $D_d{=}384$. The cross-modal scorer projects features into a shared $P{=}256$ space via bias-free projections with a scaled dot-product:
\begin{equation}
q_b = W_q t_b,\ K_b = V_b W_k^\top,\ u_{b,n} = \sigma\bigl(q_b^\top K_{b,n}/\sqrt{P}\bigr).
\label{eq:scorer}
\end{equation}
The sensitivity head $H_{\text{sens}}\!:\mathbb{R}^{D_d}\!\to\!(0,1)$ has hidden widths $384{\to}128{\to}128{\to}1$ with GELU activations~\cite{hendrycks2016gaussian}, producing $s^{\text{dino}}_{b,m}$, which is bilinearly upsampled onto the CLIP grid as $s^{\text{aligned}}_b \in (0,1)^N$. The fused score and soft mask are defined as
\begin{equation}
c_{b,n} = \alpha\, u_{b,n} - \beta\, s^{\text{aligned}}_{b,n},\quad m_{b,n} = \sigma(\gamma\, c_{b,n}),
\label{eq:combine}
\end{equation}
where $\alpha{=}\beta{=}1$ are fixed buffers and $\gamma$ is a learnable sharpness parameter initialised to $3.0$. The threshold predictor concatenates global descriptors into $x_b \in \mathbb{R}^{1795}$ and applies a $1795{\to}512{\to}512{\to}1$ MLP (GELU, dropout $0.1$):
\begin{equation}
x_b = [\, v_b^{\text{pool}};\, t_b;\, \operatorname{summ}(s^{\text{dino}}_b)\,],\quad
\hat\rho_b = \sigma\bigl(\mathrm{MLP}_\tau(x_b)\bigr),
\label{eq:predictor}
\end{equation}
where $\operatorname{summ}(s^{\text{dino}}_b) = [\,\overline{s},\,\max s,\,\tfrac{1}{256}|\{s_{b,m}{>}0.5\}|\,]$. The controller performs a safety-floored top-$k$ selection:
\begin{equation}
k_b = \max\!\bigl(k_{\min},\, \operatorname{round}((1-\rho_b)N)\bigr),\
\mathcal{S}_b = \operatorname{sort}\!\bigl(\operatorname{topk}(u_b, k_b)\bigr),
\label{eq:controller}
\end{equation}
with $k_{\min}{=}16$. The selected patches are passed through LLaVA's projector $\psi$ and concatenated with surrounding text embeddings as $E_b = [\, e^{\text{pre}}_b;\, \psi(V_b[\mathcal{S}_b,:]);\, e^{\text{post}}_b\,]$. The learnable parameters $\theta = \{W_q, W_k, H_{\text{sens}}, \mathrm{MLP}_\tau, \gamma\}$ form the DTP network, which jointly predicts $\hat\rho_b$ and $m_b$ in a single forward pass and contains $\approx 1.7$M parameters (mainly $\mathrm{MLP}_\tau$ with $\approx 1.18$M and scorer projections with $\approx 459$K). During downstream fine-tuning, only LoRA adapters~\cite{hu2022lora} are trained on the seven attention and MLP projections $\{W_q, W_k, W_v, W_o, W_{\text{gate}}, W_{\text{up}}, W_{\text{down}}\}$ in each decoder layer at rank $32$, while base LLaVA and \namet remain frozen.

%==============================================
\subsection{Training Configurations}
\label{sec:setup}
Figure~\ref{fig:sysarc}(a) summarises the data distribution across the six VQA benchmarks together with the shared LLaVA-1.5-7B backbone and the LoRA placement that is common to every configuration; Figure~\ref{fig:sysarc}(b)-(e) shows the four training paradigms evaluated in this work. All paradigms use the same backbone, LoRA configuration, and optimisation schedule, and differ only in how the model and data are divided between client and server.

\paragraph{CL} A single trainer holds the entire training set and updates the LoRA adapters via mini-batch SGD with gradient accumulation; no partitioning is performed (Figure~\ref{fig:sysarc}(b)).

\paragraph{FL} $N_C$ clients hold non-overlapping shards. Each round, the server broadcasts the global LoRA state, every client trains locally for $T_l$ micro-batches, and the server applies FedAvg~\cite{mcmahan2017communication} to aggregate the updates (Figure~\ref{fig:sysarc}(c)). Optimiser state is not shared; a single global cosine schedule is used to remain comparable to CL.

\paragraph{SL} The Llama-2-7B decoder is cut at layer $C{=}16$, with the client owning $[0, C)$ together with the \namet framework, projector, and embeddings, and the server owning $[C, L)$ together with the LM head (Figure~\ref{fig:sysarc}(d)). Within each round, $N_C$ clients train sequentially against the shared server. The cut is implemented via forward hooks on the boundary layer, preserving rotary embeddings and causal masking.

\paragraph{USL} The decoder is cut at $C_A{=}8$ and $C_B{=}24$, splitting it into a client-head, a server-body, and a client-tail (Figure~\ref{fig:sysarc}(e)). The server therefore never sees the raw image, question, answer labels, or LM-head logits; only the activations at the two cut points are server-visible. Client-head and client-tail LoRA weights are private per client, while the server-body LoRA is shared; per-client AdamW states prevent statistical leakage across clients.

%==============================================
\subsection{Robustness and Privacy Probes}
\label{sec:method-robustness-privacy}
Beyond standard VQA accuracy, we design two probes targeting robustness and privacy sensitivity. The \emph{irrelevant-question probe} pairs each test image with three categories of off-topic queries: (i) cross-image questions, where a question from one sample is paired with a different image, (ii) manually constructed nonsensical questions such as \emph{``What is the capital of France?''}, and (iii) false-premise questions referring to objects absent from the image. Responses containing refusal or uncertainty phrases from a curated set $\mathcal{R}$ (e.g., \emph{``not visible''}, \emph{``cannot determine''}) are classified as refusals; remaining non-empty responses are treated as hallucinations. We compute hallucination and refusal rates over $\mathcal{I}^{*}\subseteq\mathcal{I}$, the samples with a non-empty generation:
\begin{equation}
\scriptsize
\begin{aligned}
\mathrm{Halluc.}
&=
\frac{1}{|\mathcal{I}^{*}|}
\sum_{i\in\mathcal{I}^{*}}
\mathbb{1}\!\left[\hat{y}_i \notin \mathcal{R}\right],
\mathrm{Refusal}
&=
\frac{1}{|\mathcal{I}^{*}|}
\sum_{i\in\mathcal{I}^{*}}
\mathbb{1}\!\left[\hat{y}_i \in \mathcal{R}\right].
\end{aligned}
\end{equation}
To analyse privacy-sensitive behaviour, we additionally construct a \emph{privacy-relevant subset} per dataset by ranking test images according to the maximum DINOv2-derived sensitivity score $s_{\max}(x)=\max_p s_p(x)$ across image patches and selecting the top-$N$ samples (the sensitivity-training threshold $s_{\max}>0.5$ is met by essentially all retained samples on medical datasets). On this subset, we evaluate the sensitivity exclusion ratio:
\begin{equation}
\mathrm{ExclRatio}
=
\frac{\mathbb{E}_{p \in \mathcal{D}}\!\left[s_p\right]}
     {\mathbb{E}_{p \in \mathcal{K}}\!\left[s_p\right]},
\end{equation}
where $\mathcal{K}$ and $\mathcal{D}$ denote the sets of retained and pruned patches, respectively. Values above one indicate that the policy preferentially removes highly sensitive regions, whereas values near one suggest content-agnostic dropping. We further report closed-question exact-match accuracy and open-question recall on the same privacy-sensitive subset to assess the trade-off between privacy preservation and downstream task performance.
%%%%%%%%%%%%%%%%%%%%%%%%%%%%%%%%%%%%%%%%%%%%

%%%%%%%%%%%%%%%%% Experimental Results %%%%%%%%%%%%%
\section{Experimental Results}
\label{sec:experimental-results}
This section reports the experimental setup, then walks through the qualitative behaviour of the proposed pruning policy, the VQA accuracy of \namet across the four training paradigms and its robustness to four representative privacy attacks. Later it presents an explainability analysis of the learned policy, followed by an aggregated probe on off-topic robustness and sensitivity-aware pruning. Finally, the section concludes through a discussion of remaining limitations.

%==============================================
\subsection{Experimental Setup}
\label{sec:exp_setup}
All experiments are implemented in PyTorch on two NVIDIA RTX PRO 6000 Blackwell GPUs, benchmarking \namet against CL, FL, SL, and USL under a shared training configuration: LoRA adapters ($r{=}32$, $\alpha{=}64$, dropout $0.05$); AdamW ($\beta_1{=}0.9$, $\beta_2{=}0.999$, weight decay $0$) with peak learning rate $1{\times}10^{-4}$, cosine decay, $5\%$ warmup, and $\ell_2$ gradient clipping at $1.0$. The budget is $R{=}20$ rounds of $100$ micro-batches, applied as $100$ sequential updates in CL or distributed across $N_C{=}5$ clients with $T_l{=}20$ local steps in FL/SL/USL. We use micro-batch size $16$ with accumulation factor $2$ (effective $32$) in \texttt{bfloat16}, greedy decoding up to $20$ tokens, and early stopping (patience $5$, minimum improvement $0.005$ on validation accuracy) with the best-validation checkpoint restored for testing.

%==============================================
\subsection{Qualitative Visualisation of DTP Intermediate Signals}
\label{sec:qual_pipeline}
\begin{figure}[t]
  \centering
  \includegraphics[width=\linewidth]{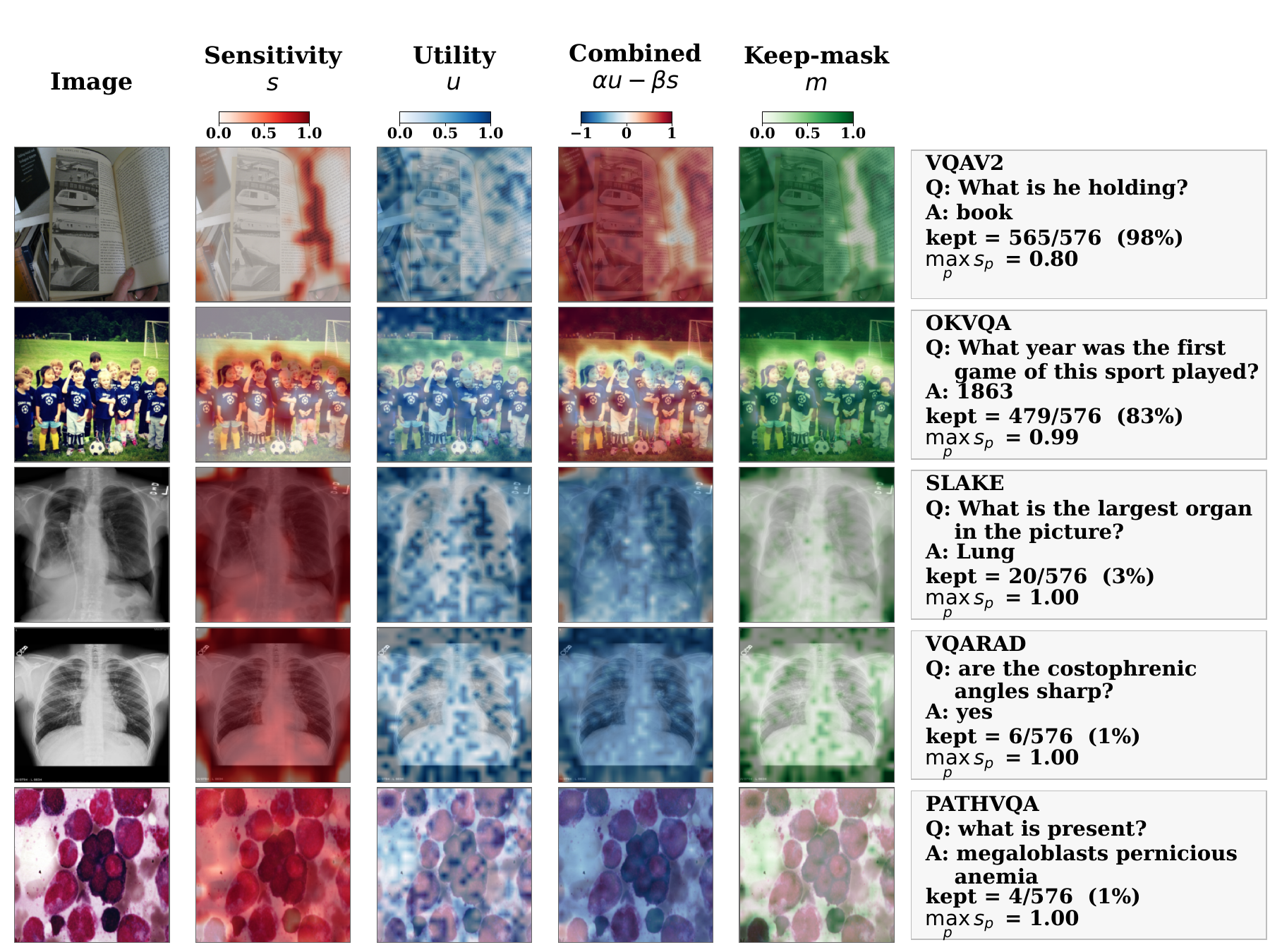}
  \vspace{-18pt}
  \caption{DTP intermediate signals (sensitivity, utility, combined score, keep-mask) on one sample per dataset.}
  \vspace{-15pt}
  \label{fig:pipeline1}
\end{figure}

Figure~\ref{fig:pipeline1} visualises four intermediate signals from one randomly selected sample in each of five VQA datasets: the per-patch sensitivity map $s$ from the DINOv2-supervised head, the question-conditioned utility score $u$, the combined score $\alpha u - \beta s$, and the final keep-mask $m$. Three patterns emerge: (i) low-sensitivity natural images (VQAv2, $\max_p s_p{=}0.80$, $565$/$576$ retained) yield diffuse distributions and high retention; (ii) privacy-sensitive natural images such as the OK-VQA example with identifiable faces ($\max_p s_p{=}0.99$, $479$/$576$ retained) produce a tighter keep-mask despite the generic question, indicating that DTP also responds to image content; (iii) medical samples (SLAKE, VQA-RAD, PathVQA, $\max_p s_p{\approx}1.00$) show concentrated sensitivity peaks and reduce the keep-mask to $4$--$20$ diagnostically relevant patches. Across all five examples, the DTP keep-mask closely follows the high-response regions of the combined score $\alpha u - \beta s$ in Figure~\ref{fig:pipeline1}, suggesting that DTP preserves the content-aware pruning behaviour while maintaining lightweight inference cost for deployment.

%==============================================
%==============================================

%==============================================
\subsection{Adaptive vs.\ Fixed Token Pruning}
\label{sec:qgtp-pruning}
\begin{figure}[t]
    \centering
    \includegraphics[width=\linewidth]{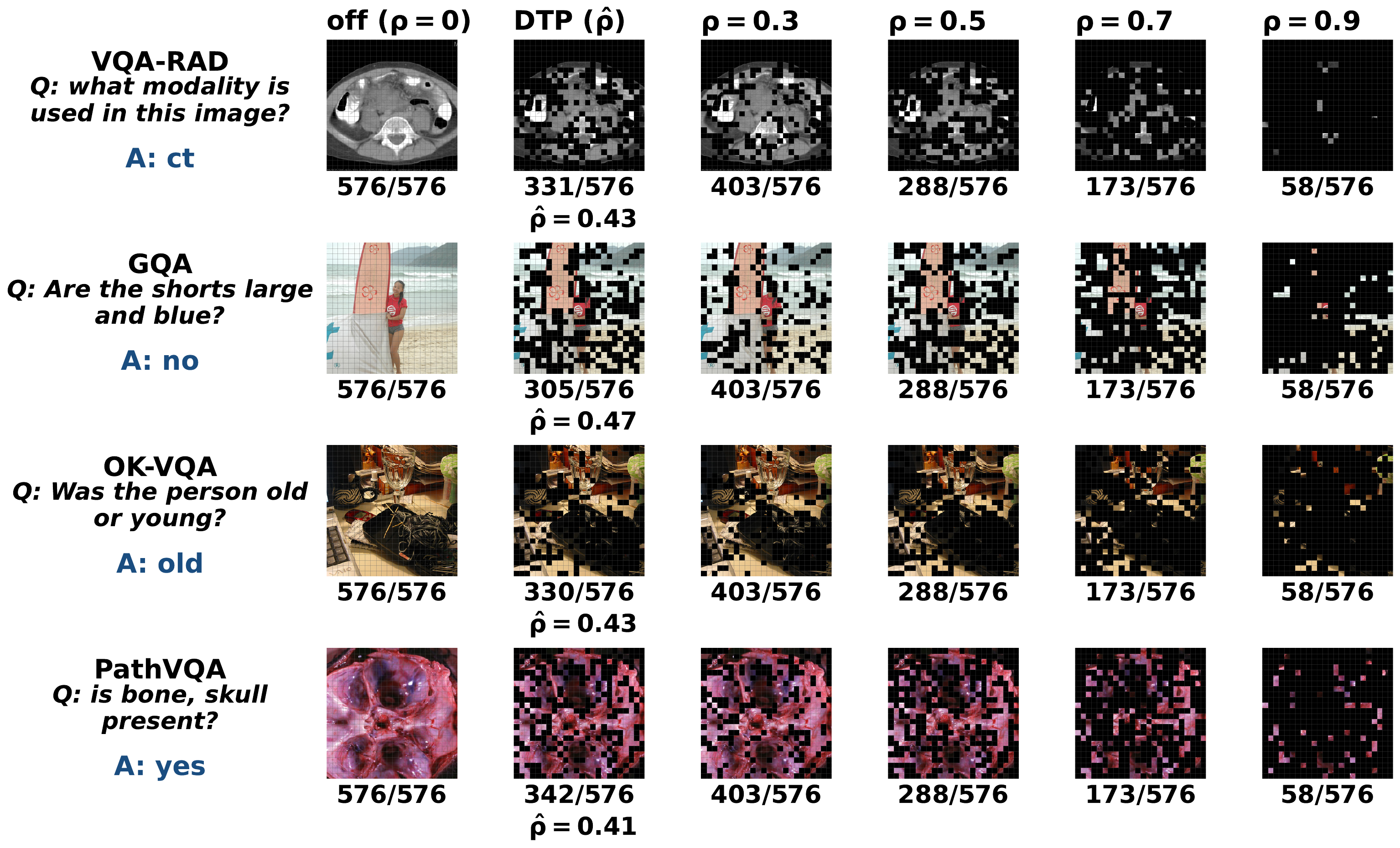}
    \vspace{-18pt}
    \caption{Question-guided token pruning under different setups.}
    \vspace{-15pt}
    \label{fig:qgtp-pruning-grid}
\end{figure}

In this part, we evaluate DTP against fixed-rate top-$k$ pruning at comparable token counts to show whether its per-sample, content-aware budget retains more semantically useful regions. For each input image, the frozen CLIP-ViT-L/14 encoder produces $N{=}576$ visual patch tokens on a $24{\times}24$ grid, and the \namet framework assigns a question-conditioned utility score to each token while predicting a per-sample pruning ratio $\hat{\rho}$ via the student head. Figure~\ref{fig:qgtp-pruning-grid} visualises the keep-masks for representative samples from VQA-RAD, GQA, OK-VQA, and PathVQA, comparing the learned DTP policy with the \texttt{off} baseline and fixed-rate settings $\rho\in\{0.3,0.5,0.7,0.9\}$, with each panel reporting the retained token count $k$ and the predicted $\hat{\rho}$. DTP retains $305$--$342$ tokens ($\hat{\rho}{\approx}0.41$--$0.47$) and consistently preserves semantically relevant regions, abdominal structures in CT, clothing in GQA, foreground subjects in OK-VQA, and tissue regions in PathVQA, whereas fixed-rate pruning removes informative regions uniformly without regard to question relevance.

%==============================================
\subsection{Accuracy-Token-Budget Trade-off over Training Paradigm}
\label{sec:summary}
Tables~\ref{tab:general_results} and~\ref{tab:medical_results} report test accuracy on three general-domain benchmarks (GQA, OKVQA, VQAv2) and three medical benchmarks (SLAKE, Path-VQA, VQA-RAD) under Direct, Federated, Split, and U-Split, with token budgets ranging from $576$ to nearly $10\%$ of the visual prefix. Both Fixed and DTP rank tokens by the same question-conditioned utility score with top-$k$ selection, differing only in budget: Fixed applies a manual ratio, whereas DTP predicts a sample-specific $\hat{\rho}_b$ via the sensitivity-conditioned $\mathrm{MLP}_\tau$. DTP is deterministic at inference for a specific sample, so each row is a single reproducible operating point since the same testing sample is used. Bold values indicate the best paradigm within each row.

On the medical benchmarks, U-Split is strongest, dominating Path-VQA and VQA-RAD across closed, open, and average accuracy, while Split is competitive on SLAKE. Fixed at $\rho{=}0.3$ ($403$ tokens) tracks the unpruned baseline within $1$--$2$ points because top-$k$ retention keeps the patches most relevant for answering. DTP instead lets the sensitivity branch dictate the budget: because medical diagnostic regions are small and sensitivity-rich, DTP allocates only $155$--$259$ tokens on these datasets. The trade-off is a few accuracy points, e.g.\ Path-VQA U-Split DTP keeps just $155$ tokens at $55.9\%$ average accuracy ($5.9$ points below Off), in exchange for the privacy-aware behaviour quantified in the attack sections.

%==============================================
%==============================================

General-domain benchmarks tell a similar story with budgets shifted upward. Although called ``natural images'', many samples carry privacy-relevant content, identifiable faces in OK-VQA, person and clothing attributes in GQA, foreground subjects in VQAv2, which DTP picks up via its DINOv2-derived sensitivity signal. The budget is sharply sample-adaptive: OK-VQA collapses to $194$ tokens ($33.7\%$, near the medical regime) due to dense face and person content, while GQA ($318$) and VQAv2 ($296$) remain more permissive. Federated and Split take turns leading the unpruned setting, U-Split dominates aggressive fixed budgets at $\rho{\in}\{0.3,0.5\}$, and DTP stays within a few points of the matched Fixed rows.

Viewed as a communication-cost study, all DTP rows lie between Fixed $\rho{=}0.5$ ($288$ tokens) and Fixed $\rho{=}0.7$ ($173$ tokens), averaging $\approx 40\%$ retention across the six datasets. DTP achieves this without manual tuning by adapting per sample: clean, low-sensitivity images keep more tokens (where the attacker had little to gain), and sensitive medical or face-heavy images are pruned harder (where the attacker had most to gain), yielding a sample-adaptive Pareto operating point that no single fixed ratio reproduces at the same accuracy.

%==============================================
\begin{table}[htpb]
\centering
\vspace{-15pt}
\caption{Final test accuracy (\%) on general-domain VQA datasets.}
\label{tab:general_results}
\footnotesize
\setlength{\tabcolsep}{4pt}
\scriptsize
\begin{tabular}{@{}ll rr cccc@{}}
\toprule
\multirow{2}{*}{\textbf{Dataset}} &
\multirow{2}{*}{\textbf{Model}} &
\multicolumn{2}{c}{\textbf{Tokens}} &
\multicolumn{4}{c}{\textbf{Accuracy (\%)}} \\
\cmidrule(lr){3-4}
\cmidrule(lr){5-8}
&
&
\textbf{Avg. Kept} &
\textbf{\%} &
\textbf{Direct} &
\textbf{Fed.} &
\textbf{Split} &
\textbf{U-Split} \\
\midrule
\multirow{6}{*}{\textbf{GQA}}
& Off     & 576 & 100.0 & 62.4 & \textbf{68.6} & \textbf{68.6} & 66.8 \\
& DTP     & 318 & 55.2  & 60.4 & \textbf{61.0} & 60.3 & 60.1 \\
& Fixed ($\rho=.3$)   & 403 & 70.0  & 59.9 & 66.8 & \textbf{69.4} & 65.0 \\
& Fixed ($\rho=.5$)   & 288 & 50.0  & 66.0 & 62.7 & 66.4 & \textbf{68.2} \\
& Fixed ($\rho=.7$)   & 173 & 30.0  & 51.1 & 57.6 & \textbf{59.0} & 56.1 \\
& Fixed ($\rho=.9$)  &  58 & 10.0  & \textbf{52.9} & 50.0 & 50.7 & 50.1 \\
\midrule
\multirow{6}{*}{\textbf{OKVQA}}
& Off     & 576 & 100.0 & 52.5 & \textbf{56.2} & 54.5 & 49.0 \\
& DTP     & 194 & 33.7  & 41.1 & \textbf{42.6} & 41.3 & 37.0 \\
& Fixed ($\rho=.3$)   & 403 & 70.0  & 50.4 & \textbf{50.7} & 50.4 & 46.3 \\
& Fixed ($\rho=.5$)   & 288 & 50.0  & 47.9 & \textbf{49.1} & 48.4 & 47.0 \\
& Fixed ($\rho=.7$)  & 173 & 30.0  & 43.1 & \textbf{44.2} & 42.3 & 37.5 \\
& Fixed ($\rho=.9$)  &  58 & 10.0  & \textbf{37.2} & 37.1 & 36.0 & 35.2 \\
\midrule
\multirow{6}{*}{\textbf{VQAv2}}
& Off     & 576 & 100.0 & 69.0 & 68.2 & \textbf{69.3} & 67.7 \\
& DTP     & 296 & 51.4  & \textbf{59.1} & 59.0 & 56.1 & 53.6 \\
& Fixed ($\rho=.3$)  & 403 & 70.0  & 66.2 & 65.0 & 66.3 & \textbf{66.6} \\
& Fixed ($\rho=.5$)  & 288 & 50.0  & 61.3 & 59.8 & 60.2 & \textbf{62.6} \\
& Fixed ($\rho=.7$)  & 173 & 30.0  & 54.3 & \textbf{54.4} & 53.6 & 53.5 \\
& Fixed ($\rho=.9$)  &  58 & 10.0  & 45.7 & \textbf{47.4} & 46.5 & 45.3 \\
\bottomrule
\end{tabular}
\vspace{-10pt}
\end{table}
%==============================================

%==============================================
\begin{table*}[!t]
\centering
\scriptsize
\caption{Closed, open, and overall accuracy (\%) on medical VQA datasets.}
\label{tab:medical_results}
\begin{tabular}{@{}ll rr ccc ccc ccc ccc@{}}
\toprule
\multirow{2}{*}{\textbf{Dataset}} &
\multirow{2}{*}{\textbf{Model}} &
\multicolumn{2}{c}{\textbf{Tokens}} &
\multicolumn{3}{c}{\textbf{Direct}} &
\multicolumn{3}{c}{\textbf{Federated}} &
\multicolumn{3}{c}{\textbf{Split}} &
\multicolumn{3}{c}{\textbf{U-Split}} \\
\cmidrule(lr){3-4}
\cmidrule(lr){5-7}
\cmidrule(lr){8-10}
\cmidrule(lr){11-13}
\cmidrule(lr){14-16}
&
&
\textbf{Avg. Kept} &
\textbf{\%} &
\textbf{Closed} &
\textbf{Open} &
\textbf{Avg} &
\textbf{Closed} &
\textbf{Open} &
\textbf{Avg} &
\textbf{Closed} &
\textbf{Open} &
\textbf{Avg} &
\textbf{Closed} &
\textbf{Open} &
\textbf{Avg} \\
\midrule
\multirow{6}{*}{\textbf{SLAKE}}
& Off
& 576 & 100.0
& 84.1 & 72.3 & 78.2
& 86.0 & 76.5 & 81.3
& \textbf{87.2} & 78.1 & \textbf{82.7}
& 85.1 & \textbf{81.1} & 82.4 \\
& DTP
& 259 & 45.0
& \textbf{79.0} & 69.8 & \textbf{74.6}
& 77.2 & 68.5 & 73.8
& 78.4 & 69.2 & 74.3
& 74.9 & \textbf{73.1} & 73.7 \\
& Fixed ($\rho=.3$)
& 403 & 70.0
& 85.3 & 74.8 & 80.2
& 83.1 & 72.9 & 78.6
& \textbf{86.0} & 75.4 & 80.7
& 83.9 & \textbf{79.4} & \textbf{80.9} \\
& Fixed ($\rho=.5$)
& 288 & 50.0
& \textbf{84.8} & \textbf{74.1} & \textbf{79.8}
& 79.2 & 67.8 & 71.0
& 82.1 & 71.4 & 75.4
& 78.3 & 71.6 & 73.8 \\
& Fixed ($\rho=.7$)
& 173 & 30.0
& 78.5 & 65.2 & 72.1
& 76.0 & 63.8 & 69.2
& \textbf{79.4} & 67.0 & 73.4
& 74.3 & \textbf{73.9} & \textbf{74.0} \\
& Fixed ($\rho=.9$)
& 58 & 10.0
& 70.2 & 58.7 & 64.5
& 68.1 & 56.5 & 62.3
& \textbf{74.5} & 61.3 & \textbf{70.2}
& 57.6 & \textbf{69.3} & 65.4 \\
\midrule
\multirow{6}{*}{\textbf{Path-VQA}}
& Off
& 576 & 100.0
& 66.5 & 50.2 & 58.4
& 65.9 & 49.6 & 58.0
& 68.8 & 52.1 & 60.5
& \textbf{70.2} & \textbf{53.4} & \textbf{61.8} \\
& DTP
& 155 & 26.9
& 57.2 & 43.0 & 50.1
& 60.4 & 47.0 & 53.7
& 62.1 & \textbf{48.7} & 55.4
& \textbf{63.5} & 48.3 & \textbf{55.9} \\
& Fixed ($\rho=.3$)
& 403 & 70.0
& 63.9 & 50.3 & 57.1
& 63.4 & 49.7 & 56.8
& 67.1 & 53.6 & 60.6
& \textbf{69.4} & \textbf{54.8} & \textbf{62.1} \\
& Fixed ($\rho=.5$)
& 288 & 50.0
& 62.1 & 48.2 & 55.1
& 61.8 & 48.0 & 55.0
& 65.5 & 51.4 & 58.5
& \textbf{67.2} & \textbf{52.6} & \textbf{59.9} \\
& Fixed ($\rho=.7$)
& 173 & 30.0
& 61.4 & 47.1 & 55.0
& 59.9 & 46.3 & 53.1
& 64.2 & 50.0 & 57.1
& \textbf{65.8} & \textbf{51.2} & \textbf{58.5} \\
& Fixed ($\rho=.9$)
& 58 & 10.0
& 60.7 & 46.5 & 55.5
& 59.8 & 45.9 & 54.6
& 63.1 & 49.2 & 56.2
& \textbf{64.7} & \textbf{50.5} & \textbf{57.6} \\
\midrule
\multirow{6}{*}{\textbf{VQA-RAD}}
& Off
& 576 & 100.0
& 62.8 & 45.7 & 54.3
& 64.0 & 47.3 & 55.7
& 60.5 & 43.8 & 52.1
& \textbf{65.3} & \textbf{48.1} & \textbf{56.7} \\
& DTP
& 221 & 38.4
& 60.4 & \textbf{44.1} & 52.9
& 57.6 & 42.0 & 50.1
& 59.8 & 43.3 & 52.4
& \textbf{61.7} & 44.0 & \textbf{53.0} \\
& Fixed ($\rho=.3$)
& 403 & 70.0
& 64.3 & 46.5 & 55.4
& 62.1 & 44.9 & 53.2
& 65.0 & 47.8 & 57.4
& \textbf{66.8} & \textbf{48.9} & \textbf{58.5} \\
& Fixed ($\rho=.5$)
& 288 & 50.0
& 61.9 & 45.0 & 54.0
& 60.5 & 43.2 & 52.9
& 56.1 & 41.8 & 48.5
& \textbf{63.4} & \textbf{46.2} & \textbf{55.4} \\
& Fixed ($\rho=.7$)
& 173 & 30.0
& 58.4 & 44.7 & 51.5
& 55.1 & 41.9 & 48.5
& 56.2 & 43.0 & 49.6
& \textbf{60.1} & \textbf{45.3} & \textbf{52.9} \\
& Fixed ($\rho=.9$)
& 58 & 10.0
& 55.0 & 44.1 & 49.6
& 50.3 & 39.8 & 45.4
& 52.0 & 41.6 & 46.8
& \textbf{56.8} & \textbf{42.7} & \textbf{49.9} \\
\bottomrule
\end{tabular}
\vspace{-10pt}
\end{table*}
%==============================================

%==============================================
\subsection{\namet Robustness Against FSHA}
\label{sec:fsha}
We next ask whether \namet limits what an adversarial split-learning server can recover from intermediate activations via FSHA~\cite{pasquini2021unleashing}. The attack targets cut layer $\ell{=}16$ of LLaVA-1.5-7B. With the victim LoRA adapters trained and frozen, we simulate FSHA by training a shadow autoencoder $(\tilde{f}, \tilde{f}^{-1})$ on $800$ public VQA-RAD images while a discriminator $D$ adversarially aligns the output distribution of $\tilde{f}$ with the real client activations ($\lambda_{\text{adv}}{=}0.2$, $\lambda_{\text{rec}}{=}1.0$). The trained inverse $\tilde{f}^{-1}$ is applied to $400$ held-out private activations to reconstruct medical images at $112{\times}112$ under \texttt{off}, fixed $\rho\in\{0.3,0.5,0.7,0.9\}$, and the learned DTP policy. 

\begin{figure}[t]
  \centering
  \includegraphics[width=\linewidth]{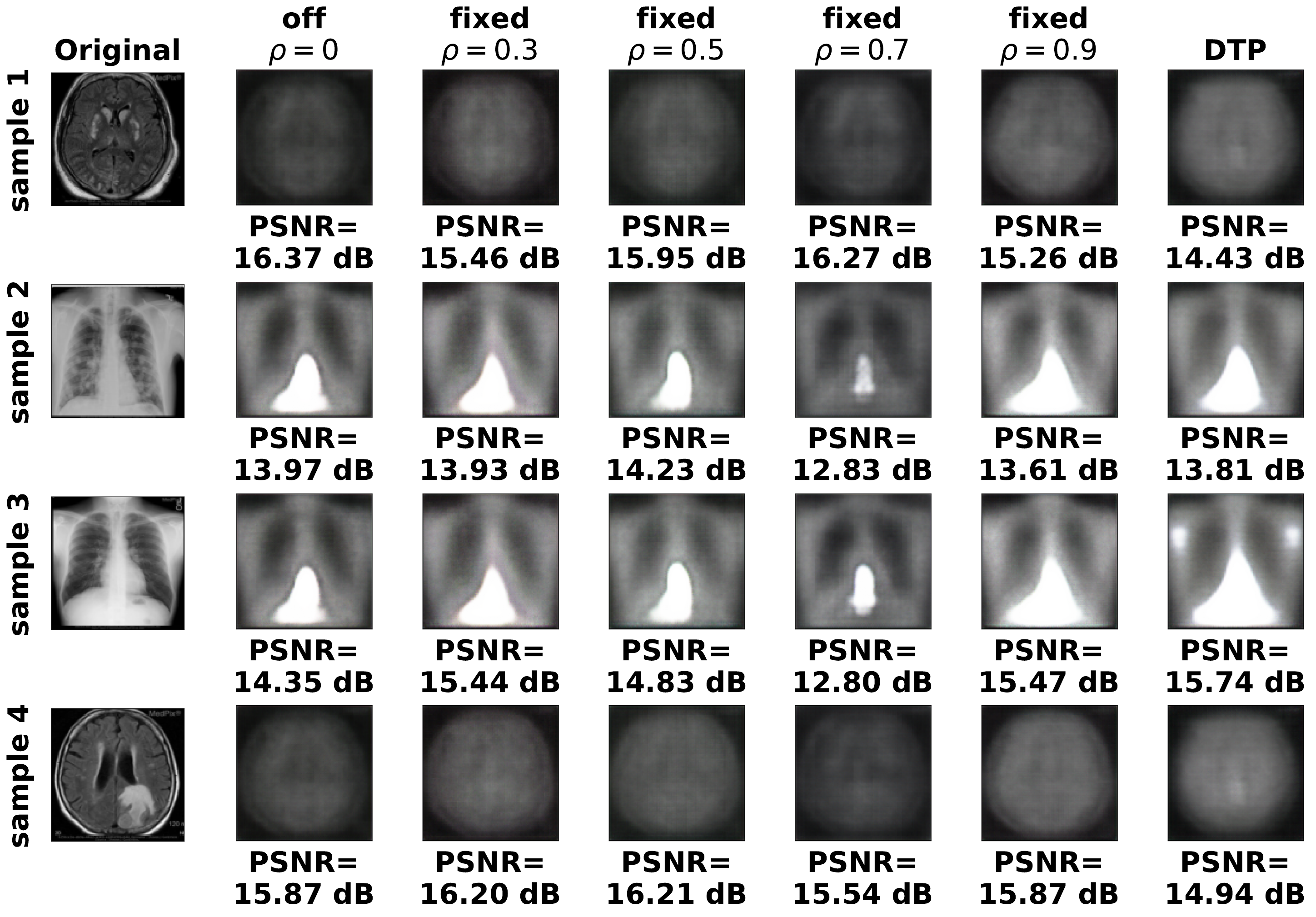}
  \vspace{-18pt}
  \caption{Qualitative FSHA reconstructions from layer-16 smashed activations under different \namet defense settings.}
  \vspace{-15pt}
  \label{fig:fsha_orig_qualitative}
\end{figure}

Figure~\ref{fig:fsha_orig_qualitative} shows that reconstructions remain low fidelity across every defense, with PSNR in $[14.0,15.2]$\,dB and cosine in $[0.86,0.88]$. Outputs degrade into coarse anatomical silhouettes, faint cranial outlines in brain MRIs or blurred thoracic regions in chest X-rays, without revealing lesion-level details, indicating that FSHA fails to recover diagnostically meaningful information from layer-$16$ representations even without pruning. The fixed setting $\rho{=}0.9$ transmits the fewest tokens ($58$ of $576$) and consequently yields the lowest reconstruction quality, while DTP attains comparable PSNR ($\approx 14.0$\,dB, cosine $\approx 0.86$) while retaining roughly $221$ tokens on VQA-RAD, placing it below the linear PSNR-vs-tokens trend.

%==============================================
%==============================================

%==============================================
\subsection{\namet Robustness Against FORA}
\label{sec:fora}
We stress-test \namet against FORA, a stronger inversion attack whose decoders are trained once on unpruned public activations and reused against every defended target, providing an upper bound on recoverable information. The setting is USL: the client keeps the embeddings and top decoder head; the malicious server controls the middle block between cuts $a{=}8$ and $b{=}24$. For each victim checkpoint we extract $800$ public and $400$ private layer-$16$ activations and train (i) a feature-space decoder targeting the $576{\times}1024$ CLIP-V grid and (ii) a pixel-space decoder targeting $112{\times}112$ RGB images (AdamW, $6$ epochs, batch size $16$, lr $10^{-4}$). We sweep $\rho\in\{0,0.3,0.5,0.7,0.9\}$ together with DTP, and compare a victim trained without the \namet framework against one co-trained with DTP on SLAKE and VQA-RAD. 

Table~\ref{tab:fora_ushape} reports pixel-space PSNR and cosine similarity; feature-space PSNR stays at or below $2.3$\,dB across all settings, indicating that layer-$16$ representations are too entangled to invert into CLIP feature space. Aggressive pruning reduces attacker PSNR by roughly $1.5$\,dB on SLAKE ($15.76$ to $14.28$) and $2.0$\,dB on VQA-RAD ($14.39$ to $12.43$) between \texttt{off} and $\rho{=}0.9$. DTP at its operating point ($\rho{=}0.42$, keeping $\sim$$335$ tokens on SLAKE; $\rho{=}0.61$, keeping $\sim$$225$ tokens on VQA-RAD) achieves the lowest PSNR in every column and sits $1.4$--$1.7$\,dB below the linearly interpolated fixed-$\rho$ trend at matched token cost. On VQA-RAD, DTP even matches the PSNR of the most aggressive fixed pruning ($\rho{=}0.9$, only $58$ tokens) while retaining roughly $4{\times}$ as many tokens, a Pareto improvement. Co-training the victim encoder with DTP adds a dataset-dependent defense: on VQA-RAD it weakens the attacker even at $\rho{=}0$ ($14.39\to 13.85$\,dB) and further at the DTP operating point ($12.42\to 12.34$\,dB), suggesting co-training shifts representations toward a harder-to-invert geometry; on SLAKE the inference-time pruning dominates and co-training has little additional effect.

%==============================================
\begin{table}[!t]
\centering
\caption{FORA reconstruction under USL.}
\label{tab:fora_ushape}
\footnotesize
\setlength{\tabcolsep}{4pt}
\scriptsize
\begin{tabular}{@{}l c cc cc cc cc@{}}
\toprule
\multirow{2}{*}{\textbf{Policy}} &
\multirow{2}{*}{$\boldsymbol{\rho}$} &
\multicolumn{4}{c}{\textbf{SLAKE}} &
\multicolumn{4}{c}{\textbf{VQA-RAD}} \\
\cmidrule(lr){3-6} \cmidrule(lr){7-10}
& &
\multicolumn{2}{c}{\textbf{w/o \namet}} &
\multicolumn{2}{c}{\textbf{co-trained}} &
\multicolumn{2}{c}{\textbf{w/o \namet}} &
\multicolumn{2}{c}{\textbf{co-trained}} \\
\cmidrule(lr){3-4} \cmidrule(lr){5-6} \cmidrule(lr){7-8} \cmidrule(lr){9-10}
& &
\textbf{PSNR} & \textbf{Cos} &
\textbf{PSNR} & \textbf{Cos} &
\textbf{PSNR} & \textbf{Cos} &
\textbf{PSNR} & \textbf{Cos} \\
\midrule
Off    & 0   & 15.76 & 0.892 & 15.76 & 0.890 & 14.39 & 0.885 & 13.85 & 0.887 \\
Fixed  & 0.3 & 15.31 & 0.878 & 15.68 & 0.885 & 14.90 & 0.878 & 14.78 & 0.879 \\
Fixed  & 0.5 & 14.94 & 0.862 & 15.39 & 0.876 & 14.40 & 0.864 & 14.46 & 0.863 \\
Fixed  & 0.7 & 14.74 & 0.852 & 15.11 & 0.868 & 13.62 & 0.845 & 13.66 & 0.842 \\
Fixed  & 0.9 & 14.28 & 0.841 & 14.55 & 0.855 & 12.43 & 0.801 & 12.36 & 0.793 \\
\midrule
DTP & 0.42 & \textbf{13.51} & \textbf{0.824} & \textbf{14.07} & \textbf{0.851} & -   & -   & -   & -   \\
DTP & 0.61 & -   & -   & -   & -   & \textbf{12.42} & \textbf{0.786} & \textbf{12.34} & \textbf{0.784} \\
\bottomrule
\end{tabular}
\vspace{-10pt}
\end{table}
%==============================================

%==============================================
\subsection{\namet Robustness Against MIA}
\label{sec:mia}
We examine whether \namet limits attribute leakage. The adversary is a malicious server observing layer-$16$ smashed activations of LLaVA-1.5-7B to infer a sensitive attribute: the binary \textsc{open}/\textsc{closed} question type on VQA-RAD, and the three-class imaging modality (CT, MRI, X-ray) on SLAKE from BoKelvin/SLAKE annotations matched by $(\text{question}, \text{answer})$. To avoid trivial length leakage, the attacker uses only the image-token portion and mean-pools it into a $4096$-dimensional vector; a frozen CLIP-T question embedding is optionally concatenated as prior knowledge. An MLP classifier is trained on $800$ public and evaluated on $400$ private samples; leakage is measured as $\mathrm{AUC}_{\text{full}}-\mathrm{AUC}_{\text{question-only}}$. We sweep all four training paradigms against six defenses: \texttt{off}, fixed $\rho\in\{0.3,0.5,0.7,0.9\}$, and DTP at $\rho{=}0.61$ on VQA-RAD and $\rho{=}0.47$ on SLAKE.

\begin{figure}[t]
  \centering
  \includegraphics[width=\linewidth]{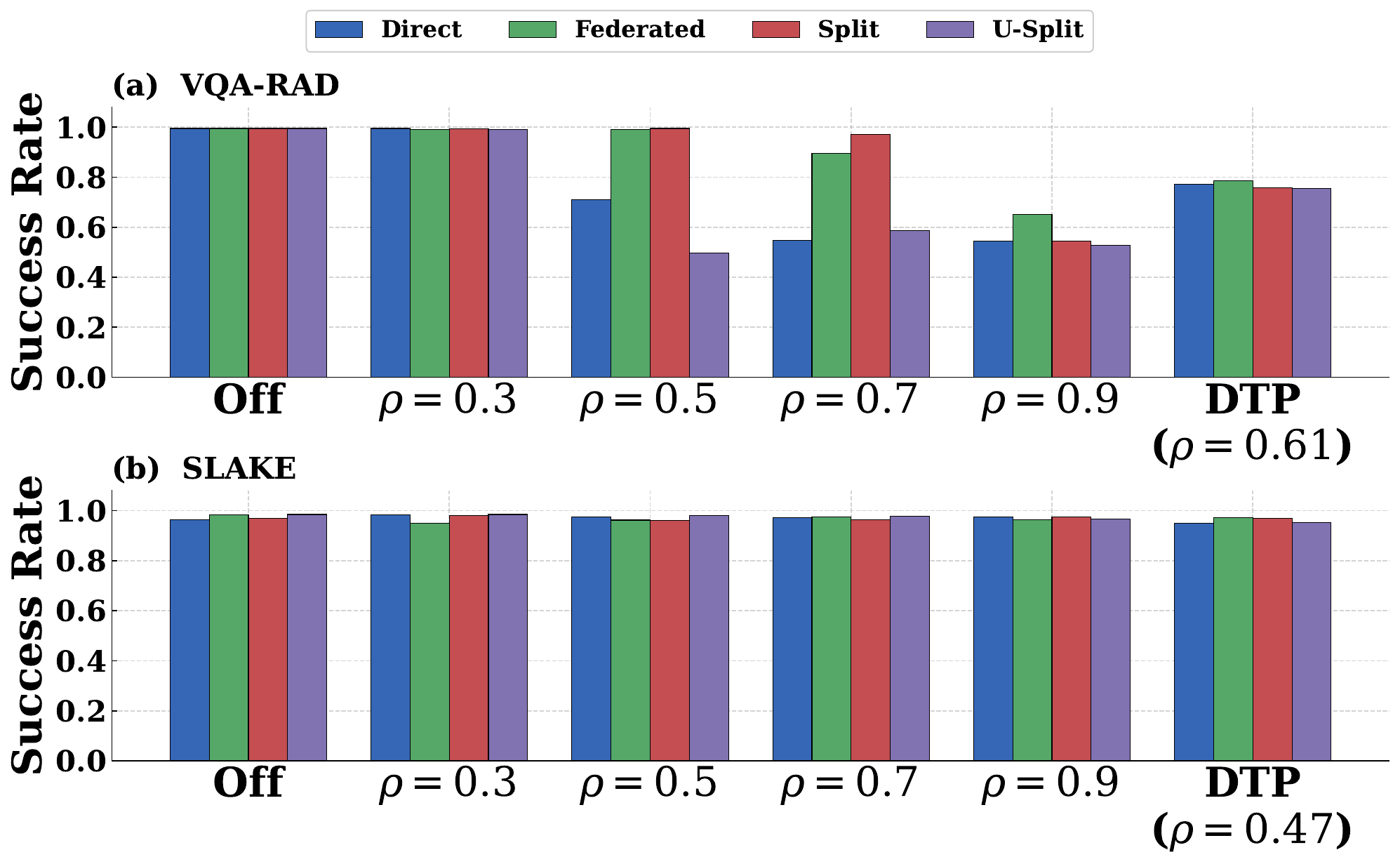}
  \vspace{-18pt}
  \caption{Attribute-inference attack accuracy across four training paradigms and six defense settings on VQA-RAD and SLAKE.}
  \vspace{-15pt}
  \label{fig:mia_acc_bars}
\end{figure}

Figure~\ref{fig:mia_acc_bars} summarises the attack. On VQA-RAD, undefended accuracy is near saturation ($\sim$$0.99$) and drops to $0.53$--$0.65$ at $\rho{=}0.9$, with Direct, Split, and U-Split bottoming near $0.55$ while Federated leaks more ($\sim$$0.65$). DTP at $\rho{=}0.61$ (retaining $\sim$$225$ tokens) sits at $0.76$--$0.79$, reducing leakage while keeping nearly four times as many tokens as the strongest fixed setting. On SLAKE, accuracy remains near saturation ($\geq 0.95$) under every defense including DTP, because the modality label is encoded in coarse image statistics (e.g.\ global intensity distributions separating MRI from X-ray) surviving heavy pruning, and because modality-specific question terms already give the attacker a strong CLIP-T prior. The persistent leakage on SLAKE reflects the low-entropy modality attribute rather than a limitation of the pruning policy.

%==============================================
%==============================================

%==============================================
\subsection{\namet Robustness Against iDLG Attack}
\label{sec:grad_leak}
\begin{figure}[t]
  \centering
  \includegraphics[width=\linewidth]{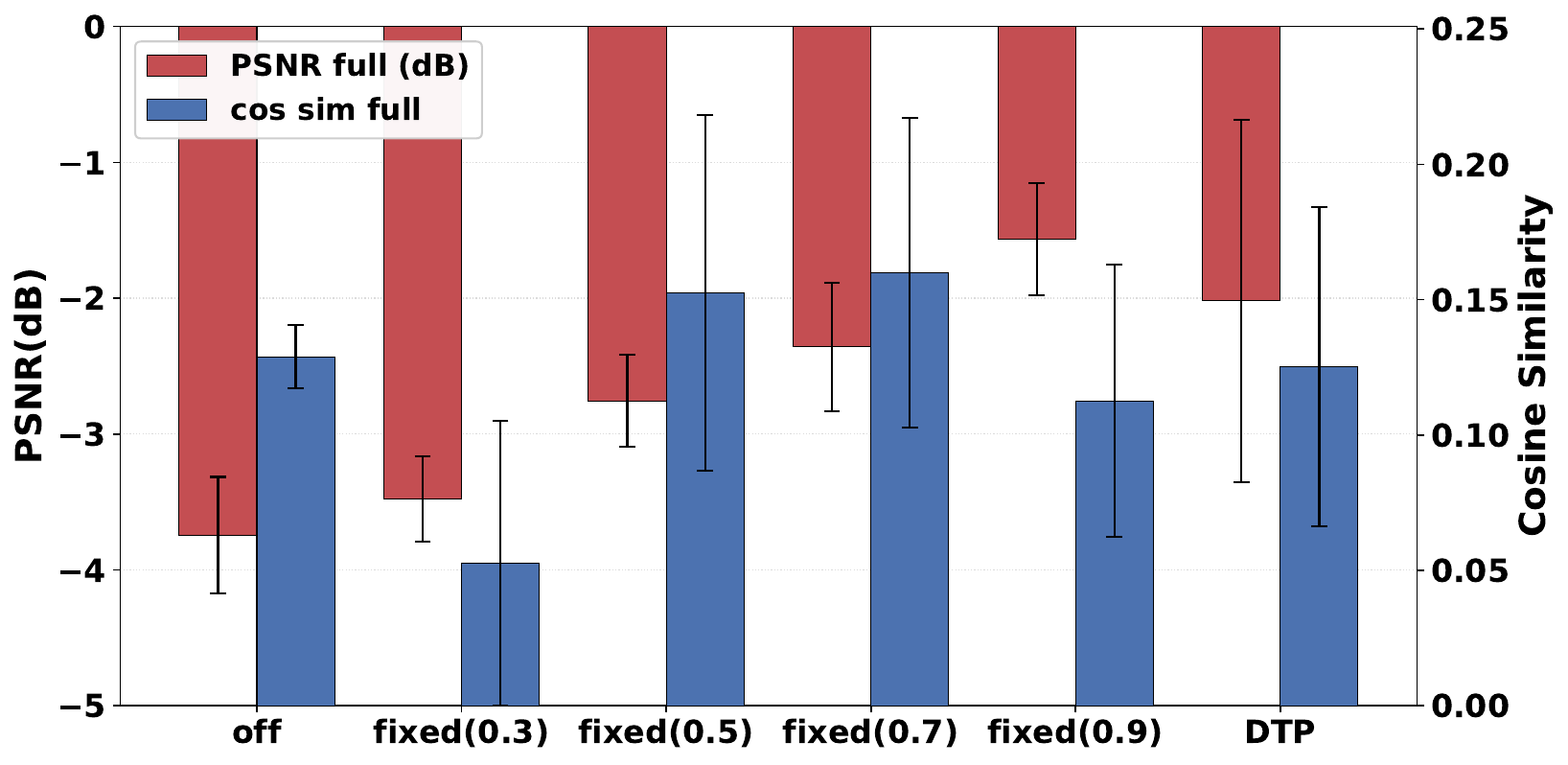}
  \vspace{-20pt}
  \caption{Gradient inversion on federated VQA-RAD.}
  \vspace{-15pt}
  \label{fig:grad_leak_fed}
\end{figure}

We test whether \namet leaks visual content through LoRA gradients during federated training. An honest-but-curious FedAvg server observes per-client LoRA gradients from one forward-backward pass over a real $(\text{image}, \text{question}, \text{answer})$ tuple and tries to reconstruct the image using DLG~\cite{zhu2019deep} and iDLG~\cite{zhao2020idlg}. The attacker initialises a dummy CLIP-V feature tensor of shape $(576, 1024)$ with Gaussian noise, applies the client's pruning policy, and optimises the dummy input with the true answer fixed, minimising $\sum_i \lVert \nabla_{\theta_i}\mathcal{L}_{\text{dummy}} - \nabla_{\theta_i}\mathcal{L}_{\text{real}}\rVert^2$ over the LoRA parameters (Adam, lr $1.0$, $300$ steps, with a small total-variation regulariser). LoRA parameters are cast to fp32 for double-backward through bf16-LLaVA. defenses include \texttt{off}, fixed $\rho\in\{0.3,0.5,0.7,0.9\}$, and DTP at $\rho{=}0.70$. Quality is measured in CLIP feature space using PSNR and cosine similarity averaged over $N{=}8$ samples. Figure~\ref{fig:grad_leak_fed} shows that gradient inversion is essentially ineffective here. Even without any defense, recovery is poor (PSNR $\approx -3.7$\,dB, cosine $\approx 0.13$), reflecting how diffusely visual information spreads across the LoRA adapters of a $7$B VLM. Pruning shifts PSNR only into $-3.7$ to $-1.6$\,dB and cosine into $0.05$--$0.16$, all well below useful image quality. DTP at $\rho{=}0.70$ lies inside this noise band ($-2.0$\,dB, $\sim$$0.13$ cosine) while retaining more tokens than fixed $\rho{=}0.9$, preserving downstream utility without weakening the (already saturated) defense.

%==============================================
\subsection{Explainability Analysis of \namet}
\label{sec:xai}
We now ask whether \namet genuinely conditions on the question rather than on generic saliency. To answer this, we visualise the DTP utility map $u \in \mathbb{R}^{24 \times 24}$ and the corresponding keep/drop decisions for a single image paired with five questions: the original GQA query, two visually grounded in-context questions, and two out-of-context (OOC) questions (arithmetic or factual queries unrelated to the image). The utility map is bilinearly upsampled and rendered with the \texttt{magma} colormap; retained patches carry a cyan contour and dropped regions are darkened. We compare DTP against the \texttt{off} baseline and fixed ratios $\rho\in\{0.3,0.5,0.7,0.9\}$, and report the retained token count $k$ and the answer from the LoRA-adapted LLaVA-1.5-7B model trained Direct on GQA.

\begin{figure}[t]
    \centering
    \includegraphics[width=0.98\linewidth]{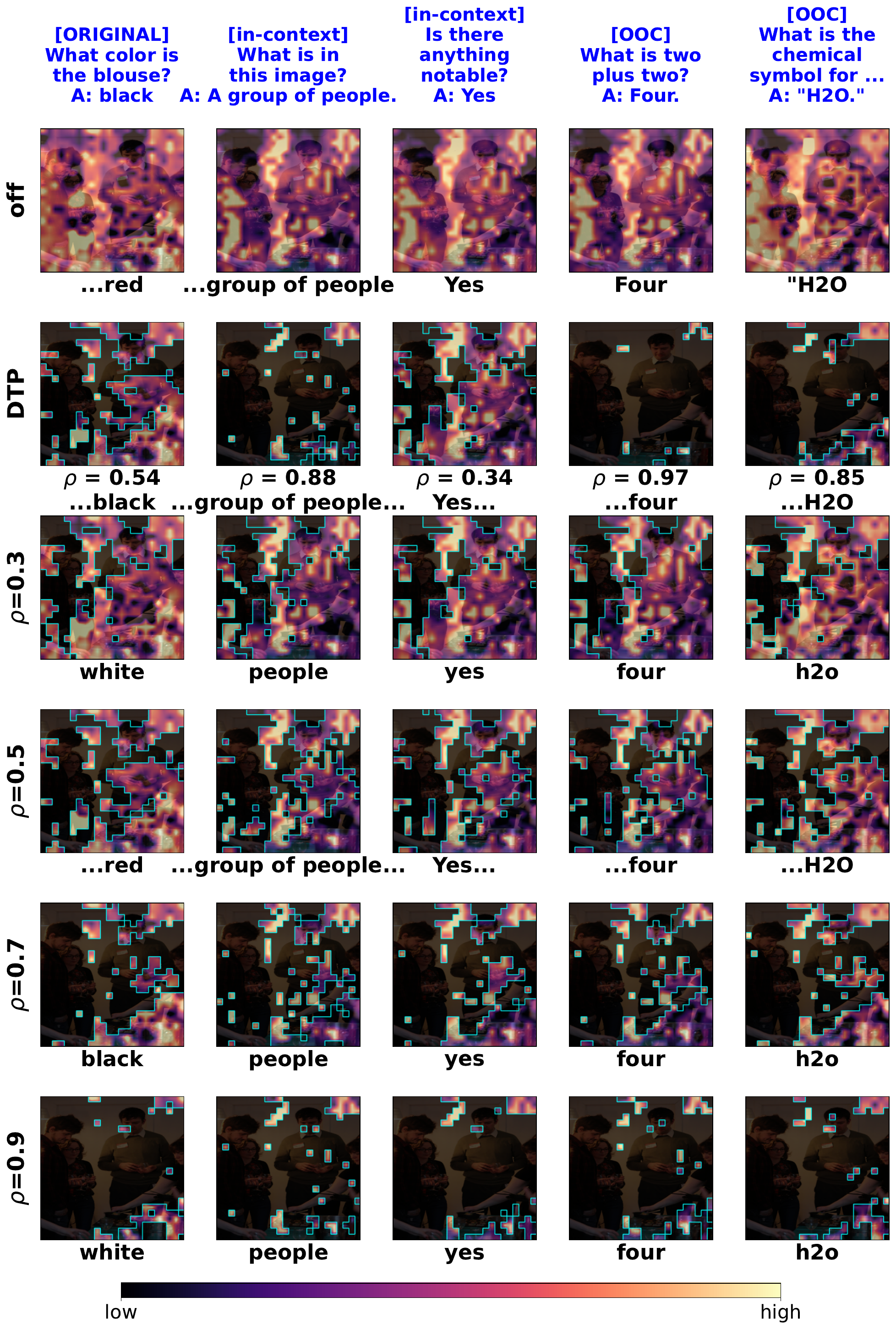}
    \vspace{-15pt}
    \caption{Question-conditioned utility maps and retained regions under multiple in-context and out-of-context questions.}
    \vspace{-15pt}
    \label{fig:xai_qcondition}
\end{figure}

%==============================================
\begin{table*}[t]
\centering
\caption{Aggregated robustness and privacy probe results across $72$ experimental dataset-setting pairs.}
\label{tab:robustness-privacy}
\small
\setlength{\tabcolsep}{4pt}
\begin{tabular}{lcccccc}
\toprule
\textbf{Mode} & \textbf{Halluc.} $\downarrow$ & \textbf{Refusal} $\uparrow$ & \textbf{Rel.\ Acc} $\uparrow$ & \textbf{Excl.\ ratio} $\uparrow$ & \textbf{Sens.\ Closed} $\uparrow$ & \textbf{Sens.\ Open} $\uparrow$ \\
\midrule
\texttt{off}        & $0.955 \pm 0.062$ & $0.045 \pm 0.062$ & $\mathbf{0.570 \pm 0.141}$ & -             & $\mathbf{0.797 \pm 0.127}$ & $\mathbf{0.468 \pm 0.215}$ \\
\texttt{DTP}        & $0.977 \pm 0.027$ & $0.023 \pm 0.027$ & $0.505 \pm 0.145$ & $\mathbf{1.201 \pm 0.177}$ & $0.736 \pm 0.155$ & $0.378 \pm 0.215$ \\
\texttt{fixed\_0.5} & $\mathbf{0.937 \pm 0.055}$ & $\mathbf{0.063 \pm 0.055}$ & $0.502 \pm 0.136$ & $0.998 \pm 0.001$ & $0.734 \pm 0.151$ & $0.420 \pm 0.209$ \\
\bottomrule
\end{tabular}
\vspace{-15pt}
\end{table*}
%==============================================
%==============================================
%==============================================
Figure~\ref{fig:xai_qcondition} shows that DTP adapts both the size and location of retained regions to question semantics. For the colour query (``What colour is the blouse?''), DTP keeps $k{=}265$ patches focused on the blouse and predicts \textit{black}, while \texttt{off} returns \textit{red}; both fixed $\rho{=}0.3$ and $\rho{=}0.9$ return \textit{white}, confirming that uniform pruning at these budgets discards blouse-relevant context. For the OOC arithmetic query (``What is two plus two?'') the policy collapses to the safety floor ($k{=}16$), signalling that no visual evidence is needed. Together these cases provide a behavioural verification that the learned policy is genuinely question-conditioned rather than a fixed saliency template.

%==============================================
%==============================================

%==============================================
\subsection{Behavioural Probes: Off-Topic Queries and Sensitivity-Aware Pruning}
\label{sec:results-robustness-privacy}
The four preceding subsections each tested resistance to a specific privacy attack. This subsection instead aggregates two behavioural probes: an off-topic robustness probe that measures how the model behaves when asked an irrelevant question, and a sensitivity-aware pruning probe that measures whether DTP preferentially drops patches that DINOv2 marks as privacy-relevant. Table~\ref{tab:robustness-privacy} summarises both probes aggregated over all $72$ dataset $\times$ defense combinations. For the off-topic probe, hallucination rates stay high ($0.94$--$0.98$) and refusal rates low ($0.02$--$0.06$) under every pruning mode, confirming that the tendency to answer nonsensical questions is set primarily by the underlying language model rather than by visual token pruning. Relevant-question accuracy is essentially flat ($0.50$--$0.57$), meaning aggressive pruning at $\rho{=}0.5$ or the DTP policy preserves most of the task-relevant features. On the sensitivity-aware pruning probe, DTP achieves an exclusion ratio of $1.20 \pm 0.18$, meaning dropped patches carry on average about $20\%$ higher sensitivity than retained ones, whereas fixed-rate pruning at $\rho{=}0.5$ stays at $0.998$, indistinguishable from content-agnostic dropping. The cost is a moderate utility hit on the same sensitive subset (closed accuracy $0.797\to 0.736$, open recall $0.468\to 0.378$), an expected trade-off because privacy-rich regions overlap with the most informative parts of the image. Overall, DTP gives targeted suppression of sensitive content while leaving off-topic robustness untouched.

%==============================================
\subsection{Discussion and Limitations}
\label{sec:discussion}
The four research questions can now be addressed. For \textbf{RQ1}, DTP compresses the visual prefix to nearly $40\%$ of its original size while reducing the effectiveness of all evaluated attack families, including FSHA, FORA, MIA, and iDLG. Despite this reduction, VQA performance remains close to the unpruned baseline across CL, FL, SL, and USL settings. These results demonstrate that communication efficiency and privacy preservation can be improved jointly without causing significant degradation in task accuracy. For \textbf{RQ2}, the dynamic threshold predictor consistently outperforms both fixed-ratio and training-free pruning approaches on the privacy-utility trade-off. Under the same token budget, it achieves $1.4$--$1.7$\,dB lower FORA reconstruction quality than interpolated fixed-$\rho$ pruning, and on VQA-RAD it reduces MIA accuracy by nearly half at operating points unattainable by any single fixed ratio. For \textbf{RQ3}, the observed sensitivity exclusion ratio of $1.20 \pm 0.18$ (compared to approximately $1.00$ for fixed pruning), together with stronger pruning behaviour on face-rich and medical images, suggests that DTP preferentially suppresses patches that are privacy-sensitive and easier to reconstruct instead of applying uniform pruning. For \textbf{RQ4}, the explainability analysis in Section~\ref{sec:xai} further confirms this behaviour. The policy collapses to the safety floor for out-of-context questions while preserving question-relevant regions for in-context queries, consistent with the optimisation objective.
 
Several limitations remain. First, low-entropy attributes such as imaging modality still leak comparatively heavily even under aggressive pruning, since coarse global statistics and modality-specific question terms may already expose the target before visual features are transmitted. Second, the experiments are limited to LLaVA-1.5-7B with fixed cut-layer configurations (SL at $\ell{=}16$ and USL at $a{=}8$, $b{=}24$), meaning that the resulting privacy-utility trade-offs may vary for deeper cuts, larger backbones, or different fine-tuning strategies. Third, DTP relies on a fixed DINOv2-based sensitivity model during training and may generalise poorly against adaptive attackers that explicitly exploit its token-retention behaviour.
%%%%%%%%%%%%%%%%%%%%%%%%%%%%%%%%%%%%%%%%%%%%

%%%%%%%%%%%%%%%%% Conclusion %%%%%%%%%%%%%
\section{Conclusion}
\label{sec:conclusion}
We presented \namet, a question-guided and privacy-aware token-pruning framework for distributed VLM-based VQA, where DTP jointly determines both the number and selection of CLIP visual tokens transmitted across the cut layer before reaching a LoRA-adapted LLaVA-1.5-7B backbone. Across CL, FL, SL, and USL settings on six general-domain and medical VQA benchmarks, DTP achieves a strong privacy-utility trade-off, matching or outperforming the best fixed-ratio defenses against FSHA, FORA, DLG/iDLG, and attribute-inference MIA while preserving substantially more tokens than aggressive uniform pruning approaches. However, token pruning alone is insufficient to guarantee formal privacy protection. Future work will explore extending DTP with alternative mechanisms, adversarial co-training against adaptive attackers, broader generalisation across VLM architectures and modalities such as 3D medical volumes and video, and integration with differential privacy and secure aggregation to transform the empirical protection demonstrated here into formal guarantees.
%%%%%%%%%%%%%%%%%%%%%%%%%%%%%%%%%%%%%%%%%%%%

%%%%%%%%%%%%%%%%% LLM Usage %%%%%%%%%%%%%
\section{LLM Usage Statement}
\label{sec:llm_usage}
Large language models (LLMs) were used during preparation of this manuscript as assistive tools for writing and code debugging. All AI-assisted outputs were reviewed, verified, and corrected by the authors, who take full responsibility for the correctness, originality, and integrity of the work. We detail the role of LLMs below in accordance with the three criteria of originality, transparency, and responsibility-

\paragraph{Originality}
LLMs were used for editorial purposes, and all outputs were inspected by the authors to ensure accuracy and originality. LLMs assisted in drafting and refining portions of the text, while every generated passage was carefully reviewed, fact-checked, and revised by the authors to align with the underlying research findings. All figures, tables, and quantitative results were produced entirely by the authors without AI assistance. The authors remain fully responsible for the content of the paper, including all written text, technical descriptions, and reported numerical values.
\paragraph{Transparency}
LLMs were also used during code development, primarily for debugging and identifying implementation issues. However, the experimental code required substantial manual design, step-by-step correction, and multiple iterations of testing before reaching a working state, as AI-generated suggestions alone were insufficient to produce the final pipeline. Every result reported in this manuscript was obtained by running the author-verified code, and all outputs were validated against expected behaviour before inclusion. LLMs were not used to design the core methodology, model architecture, or experimental protocol of \namet, which were conceived and validated by the authors.
\paragraph{Responsibility}
LLMs were used responsibly throughout the preparation of this work. No sensitive, private, or proprietary information, including dataset content, or participant data was shared with any AI tool during writing or debugging. The authors ensured that all interactions with LLMs respected ethical considerations regarding data ownership and intellectual property. The use of LLMs was limited to general-purpose writing assistance and code debugging, and did not influence the scientific contributions or claims.
%%%%%%%%%%%%%%%%%%%%%%%%%%%%%%%%%%%%%%%%%%%%

% \bibliographystyle{unsrt}
\bibliographystyle{IEEEtran}
\bibliography{References}

%%%%%%%%%%%%%%%%%%%%%%%%%%%%%%
%% Appendix %%%%%%%%%%%%%%%%%%
\appendix

\subsection{Comparison with Prior Work}
\label{sec:related-comp}

Table~\ref{tab:related} compares \namet with recent approaches from two separate research directions: visual-token pruning for LLaVA-style VLMs (FastV~\cite{chen2024image}, SparseVLM~\cite{zhang2024sparsevlm}, PyramidDrop~\cite{xing2024pyramiddrop}, VisionZip~\cite{yang2025visionzip}) and representation-level privacy protection for split or split-federated learning (NoPeek~\cite{vepakomma2020nopeek}). The token-pruning methods focus on inference efficiency in centralized settings and do not evaluate robustness against feature-reconstruction or membership-inference attacks. NoPeek and ResSFL are designed for small CNN-based models on datasets such as CIFAR and CelebA, have not been extended to VLMs, and NoPeek remains vulnerable to FSHA attacks~\cite{pasquini2021unleashing}. To the best of our knowledge, no prior work jointly integrates question-guided token pruning and privacy-aware defence within a distributed VLM framework. Under the closest comparable setting on GQA using LLaVA-1.5-7B, \namet achieves $60.4\%$ accuracy with $318$ retained tokens in Direct mode, comparable to VisionZip ($59.3\%$) and PyramidDrop ($60.1\%$) using $192$ tokens. The higher token budget reflects the sensitivity-driven nature of DTP, which prioritizes privacy alongside efficiency. Beyond utility, \namet reduces FORA reconstruction PSNR by $1.4$--$1.7$\,dB on SLAKE and VQA-RAD and decreases VQA-RAD membership-inference accuracy from $0.99$ to $0.76$, metrics not reported by existing token-pruning approaches.

\begin{table}[ht]
\centering
\caption{Comparison of \namet with other related works.}
\label{tab:related}
\setlength{\tabcolsep}{3pt}
\begin{tabular}{@{}l c c c c c@{}}
\toprule
\textbf{Method} & \textbf{Venue} & \textbf{Backbone} & \textbf{Q-guided} & \textbf{Sens-aware} & \textbf{GQA} \\
\midrule
FastV~\cite{chen2024image}             & ECCV'24  & LLaVA-1.5 & \checkmark & ---       & $52.7$ \\
SparseVLM~\cite{zhang2024sparsevlm}    & ICML'25  & LLaVA-1.5 & \checkmark & ---      & $57.6$ \\
VisionZip~\cite{yang2025visionzip}     & CVPR'25  & LLaVA-1.5 & ---        & ---        & $59.3$ \\
PyramidDrop~\cite{xing2024pyramiddrop} & CVPR'25  & LLaVA-1.5 & \checkmark & ---       & $60.1$ \\
\midrule
\textbf{\namet (Ours)}                 & ---      & LLaVA-1.5 & \checkmark & \checkmark  & $\mathbf{60.4}$ \\
\bottomrule
\end{tabular}
\vspace{-10pt}
\end{table}

\subsection{Ablation Study}
\label{subsec:ablation_pareto}
\begin{figure}[t]
    \centering
    \includegraphics[width=0.95\linewidth]{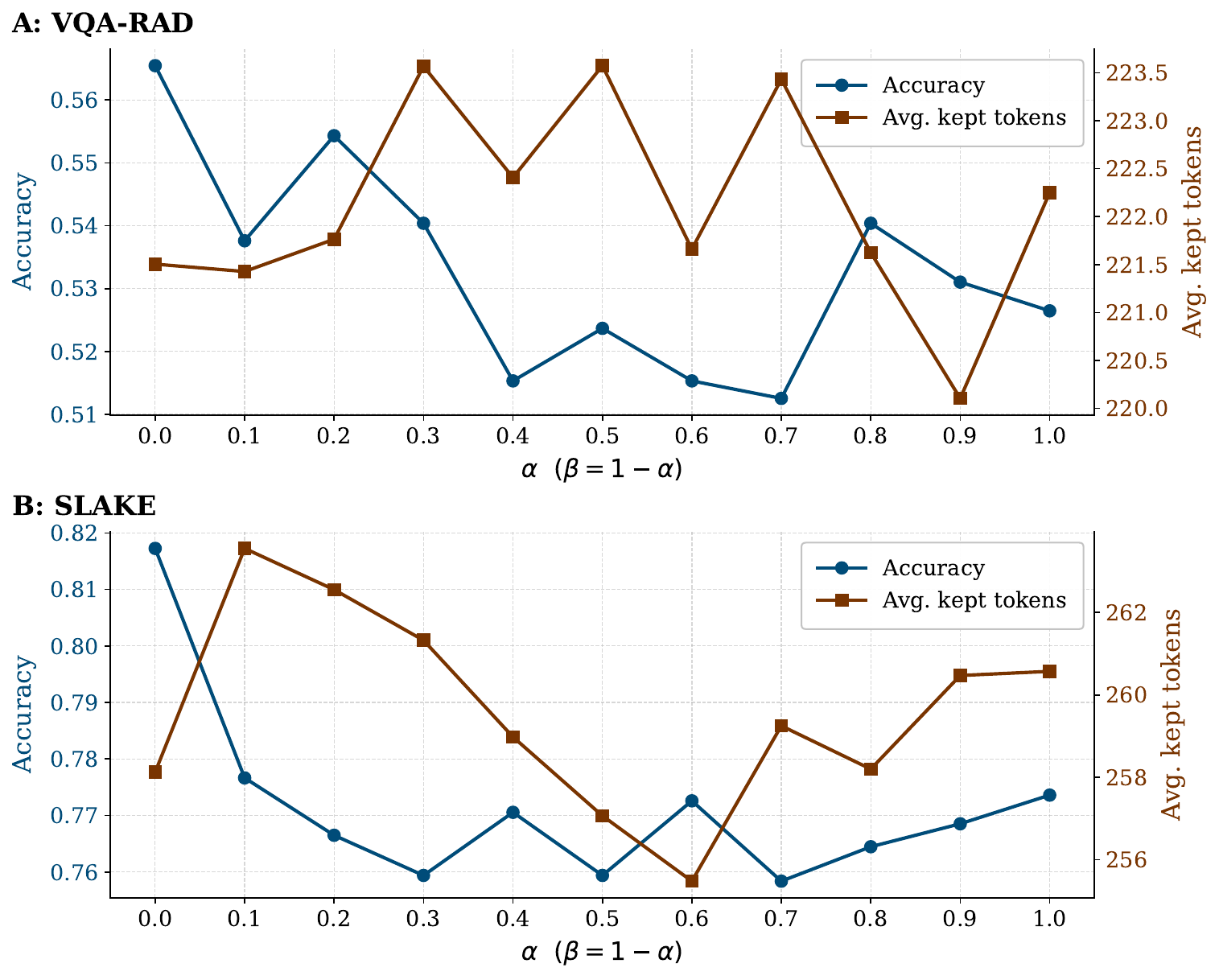}
    \caption{Ablation study of the utility-privacy trade-off.}
    \label{fig:pareto_ablation}
\end{figure}

To analyse the influence of the utility weight $\alpha$ and privacy weight $\beta$ on the proposed dynamic threshold predictor, we sweep the simplex $\alpha \in \{0.0, 0.1, \ldots, 1.0\}$ while enforcing $\beta = 1 - \alpha$, resulting in eleven student models. For each $(\alpha, \beta)$, the student is retrained with teacher mask targets recomputed under the same weighting, ensuring consistency between the distillation objective and the deployment-time scoring function $\alpha\,\mathbf{u} - \beta\,\mathbf{s}$. Each trained student is then integrated into the QGTP controller and used for LoRA fine-tuning of LLaVA-1.5-7B on VQA-RAD and SLAKE under identical schedules, learning rates, and early-stopping settings, isolating the effect of $(\alpha, \beta)$. We report test accuracy and the average number of retained visual tokens per image as complementary measures of task performance and information exposure. The resulting Pareto trade-off is illustrated in Figure~\ref{fig:pareto_ablation}.
Figure~\ref{fig:pareto_ablation} highlights two observations. First, the best performance on both datasets is achieved at $\alpha = 0$, where token ranking is determined entirely by the privacy-sensitivity signal $\beta\,\mathbf{s}$. Under this setting, VQA-RAD achieves an accuracy of $0.566$ and SLAKE reaches $0.817$, both outperforming the pure-utility configuration at $\alpha = 1$ ($0.527$ and $0.774$, respectively). This suggests that sensitivity-aware token selection naturally preserves the most diagnostically relevant regions while discarding irrelevant or noisy content that utility-only ranking tends to retain. Second, the average number of retained tokens remains nearly constant across the sweep, varying only from approximately $220$--$224$ on VQA-RAD and $256$--$262$ on SLAKE. This stability indicates that $\hat{\rho}$ is controlled mainly by the global predictor inputs $(\mathbf{c}_{\text{pool}}, \mathbf{t}, \mathbf{s}_{\text{summary}})$ rather than directly by $(\alpha, \beta)$. Consequently, the trade-off is driven not by retaining more or fewer tokens but by the criterion used to rank them, with the privacy-aware ranking proving most effective for medical VQA performance.

\end{document}